\documentclass{article}

    \PassOptionsToPackage{numbers, compress}{natbib}

    \usepackage[final]{neurips_2025}

\usepackage[utf8]{inputenc} 
\usepackage[T1]{fontenc}    

\usepackage{url}            
\usepackage{booktabs}       
\usepackage{amsfonts}       
\usepackage{amsmath}
\usepackage{nicefrac}       
\usepackage{microtype}      
\usepackage{xcolor}         
\usepackage{subcaption}
\usepackage{graphicx}
\usepackage{algorithm}
\usepackage{amssymb,amsmath,amsthm,enumerate,comment,mathtools,thmtools}
\usepackage{hyperref}  
\usepackage{thm-restate}
\usepackage[capitalize]{cleveref}
\usepackage{tabularx}
\usepackage{algorithm}
\usepackage{algorithmic}

\title{LLM-Driven Composite Neural Architecture Search for Multi-Source RL State Encoding}

\author{
    Yu Yu \\
    Shanghai Jiao Tong University
    \And 
    Qian Xie\thanks{Correspondence to: Qian Xie <\textsc{qx66@cornell.edu}>.} \\
    Cornell University
    \And Nairen Cao \\ New York University
    \And Li Jin\thanks{Correspondence to: Li Jin <\textsc{li.jin@sjtu.edu.cn}>.} \\
    Shanghai Jiao Tong University
}
\date{}

\graphicspath{{images/}}

\begin{document}

\maketitle

\begin{abstract}
Designing state encoders for reinforcement learning (RL) with multiple information sources---such as sensor measurements, time-series signals, image observations, and textual instructions---remains underexplored and often requires manual design.
We formalize this challenge as a problem of composite neural architecture search (NAS), where multiple source-specific modules and a fusion module are jointly optimized.
Existing NAS methods overlook useful side information from the intermediate outputs of these modules---such as their representation quality---limiting sample efficiency in multi-source RL settings.
To address this, we propose an LLM-driven NAS pipeline in which the LLM serves as a neural architecture design agent, leveraging language-model priors and intermediate-output signals to guide sample-efficient search for high-performing composite state encoders.
On a mixed-autonomy traffic control task, our approach discovers higher-performing architectures with fewer candidate evaluations than traditional NAS baselines and the LLM-based GENIUS framework.
\end{abstract}

\section{Introduction}
Reinforcement learning (RL) often requires transforming raw observations into compact state representations.
In many real‐world domains, the state is observed through multiple heterogeneous sources---such as sensors, time‐series signals, images, or text---necessitating specialized encoders for each source and a fusion module to combine them.
Existing RL systems typically rely on \emph{manually} designed encoders, which can be suboptimal and hard to generalize across environments.
While neural architecture search (NAS) offers a way to automate encoder design, most NAS methods target single‐modality supervised tasks and overlook the composite, multi‐source nature of RL state encoding.
Moreover, evaluating architectures in RL is notoriously costly, as each candidate must be trained through thousands of simulator interactions, making sample efficiency a key challenge.

We formulate the underexplored problem of \emph{composite NAS} for RL state encoding, where multiple source‐specific modules and a fusion module are jointly optimized.
Beyond sample efficiency, a further challenge is leveraging \emph{side information}---such as the representation quality or specialization of each module---signals that existing NAS methods typically ignore or cannot model effectively.

To address this limitation, we propose an LLM-driven NAS pipeline that leverages language-model priors over module design choices and their representation quality to automatically discover high-performing composite state encoder architectures for RL. Our main contributions are:
\begin{enumerate}
    \item We introduce and formally define the problem of composite NAS for state encoding in RL with multiple information sources.
    \item We propose an LLM-driven NAS pipeline that incorporates language-model priors to guide the search using side information about module representation quality.
    \item We instantiate and evaluate the proposed framework on a mixed-autonomy traffic control task, demonstrating improved search efficiency and RL performance.
\end{enumerate}

\section{Background and Related Works}
\subsection{RL state encoding}
State encoding is a form of representation learning that maps raw observations (e.g., images, textual descriptions, or sensor measurements) into compact latent representations that can be used by RL agents as input for policy and value estimation.
Typical architectures include convolutional neural networks (CNNs) for image observations, Transformer or recurrent networks (e.g., LSTMs, GRUs) for textual or time-series inputs, and feed-forward networks (FFNs) for structured inputs.

Various strategies have been explored for obtaining such representations. A widely adopted approach is \emph{end-to-end training}, where the encoder is jointly optimized with the RL policy using algorithms such as Proximal Policy Optimization (PPO), optimizing rewards directly. Alternatively, some works pretrain the encoder using self-supervised or contrastive learning objectives (e.g., representation consistency across views) before fine-tuning in RL.
In this work, we focus on the end-to-end setting, where the encoder architecture is optimized jointly with the RL agent.

\subsection{Neural architecture search}
Neural architecture search (NAS) aims to automatically discover high-performing neural architectures.
For RL, NAS methods can be applied either in conjunction with end-to-end training---searching for architectures that directly maximize task rewards---or in a two-stage manner where the architectures are optimized for auxiliary objectives such as contrastive loss, and then transferred to RL. In this work, we focus on the former, i.e., searching architectures trained end-to-end with the RL agent.

\paragraph{Traditional methods.}
A wide range of NAS algorithms have been developed, including gradient-based DARTS \citep{liu2018darts}, RL-based ENAS \citep{pham2018efficient}, and evolutionary-based PEPNAS \citep{xue2024neural}, and Bayesian optimization, including Gaussian process–based methods such as those implemented in BoTorch \citep{balandat2020botorch} with mixed-type (ordinal and categorical) kernels, as well as BOHB \citep{falkner2018bohb} and BANANAS \citep{white2021bananas}.

\paragraph{LLM-based methods.}
Recently, LLMs have been used to guide architecture search by generating architecture descriptions or candidates to discover high-performing architectures. Representative methods include GENIUS \citep{zheng2023can}, LLMatic \citep{nasir2024llmatic}, LAPT-NAS \citep{zhou2025design}, and SEKI \citep{cai2025seki}. However, they are primarily designed for single-modality supervised learning tasks (e.g., image classification) and do not consider composite modules or RL state encoding.

\subsection{LLM for RL}
\citet{yan2024efficient} investigates the use of LLMs as action priors to guide policy learning in RL, but not as neural architecture priors for RL state encoding. 

Recent surveys provide broader perspectives: \citet{schoepp2025evolving} categorizes three roles of LLMs in RL---Agent, Planner, and Reward---and further discusses modifying LLM architectures to serve directly as state representations, while \citet{cao2024survey} categorizes LLMs into four roles---information processor, reward designer, decision-maker, and generator---with representation learning discussed under the generator role. 

Closer to our goal of improving state representations, LESR (LLM-Empowered State Representation)~\citep{wang2024llm} uses LLMs to directly produce structured state encodings for downstream RL agents, demonstrating gains in representation quality and sample efficiency.
In contrast, our work adopts a complementary perspective: we employ LLMs to guide neural architecture search for state encoders, where the resulting architectures---rather than the LLM itself---constitute the state representation.

\section{LACER: An LLM → State Encoder → RL Pipeline}
Our approach, \textbf{L}LM-driven Neural \textbf{A}rchitecture Search for \textbf{C}omposite State \textbf{E}ncoders in \textbf{R}L (LACER) iteratively uses an LLM to generate candidate state encoder architectures, evaluates each candidate in an RL environment, and feeds the resulting performance back to the LLM.

\begin{figure}[t!]
    \centering
    \includegraphics[width=\textwidth]{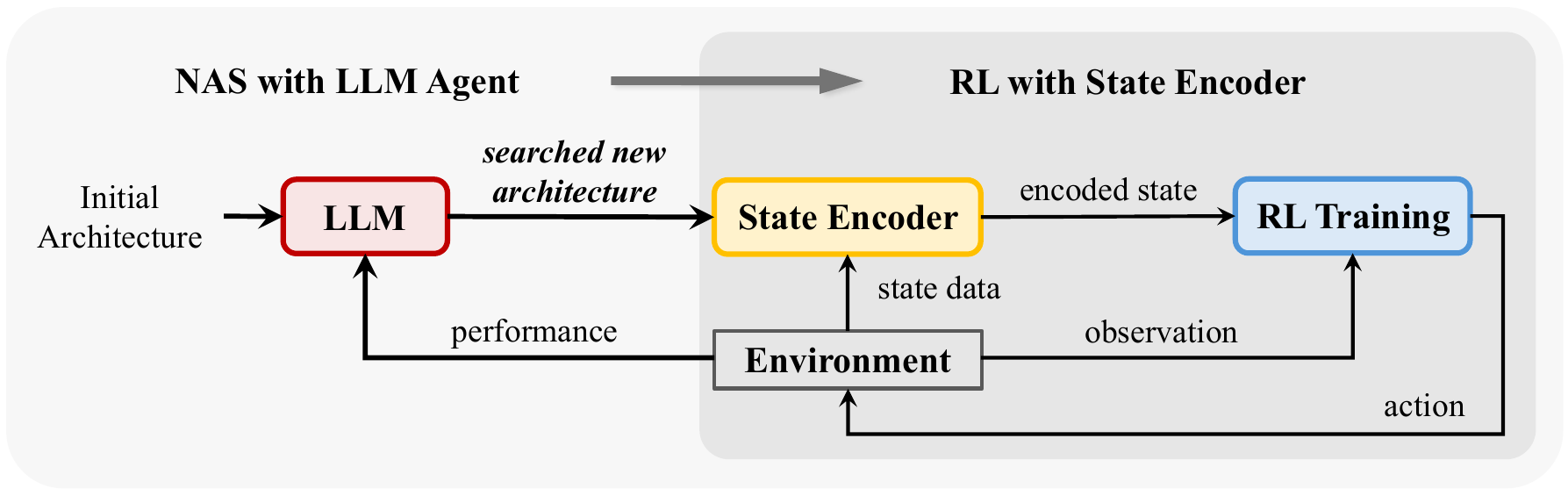}
    \caption{Our LLM design agent → architecture generation → RL training \& evaluation pipeline. In this pipeline, the LLM serves as a neural architecture design agent that perceives performance signals, reasons over conversation history, and decides on candidate architectures, facilitating iterative refinement via closed-loop interaction.}
    \label{fig:pipeline}
\end{figure}

\subsection{Problem Setup}
\label{sec:problem_setup}
We consider an RL agent that interacts with an environment and receives observations composed of multiple input sources, such as sensor values, time-series signals, textual instructions, or image observations. Each input source may require a different type of neural architecture (e.g., MLP/FFN, Transformer, CNN, RNN) to extract relevant features. Instead of searching for a single shared encoder, our goal is to automatically discover a set of architecture modules---one for each input source---and a fusion module that combines their outputs into a final latent state representation.

Formally, let $x = (x_1, \dots, x_M)$ denote the raw observations from $M$ input sources, and define the overall state encoder as
\[
s = g_{\phi}\bigl(f_{\theta_1}(x_1), \dots, f_{\theta_M}(x_M)\bigr),
\]
where each $f_{\theta_i}$ is a neural module for source $i$ with design choices $\theta_i$, and $g_{\phi}$ is a fusion module with design choices $\phi$. 
Let $\mathcal{E}$ denote the downstream RL environment. 
Our goal is to search over design choices $\{\theta_i\}_{i=1}^{M}$ and $\phi$ to maximize
\[
\mathcal{M}\left(\pi \circ g_{\phi} \circ (f_{\theta_1}, \dots, f_{\theta_M}); \mathcal{E}\right),
\]
where $\mathcal{M}(\cdot; \mathcal{E})$ measures the performance of the RL policy $\pi$ in environment $\mathcal{E}$ (e.g., average traffic speed or average return), which may differ from the reward used to train $\pi$.

Equivalently, varying $\theta_i$ selects a specific function $f_{\theta_i}$ from the family 
$\mathcal{F}_i = \{f_{\theta_i} \mid \theta_i \in \Theta_i\}$, 
and the search therefore operates over the Cartesian product 
$\Theta_1 \times \cdots \times \Theta_M \times \Phi$ of design choices.

Unlike function network with partial evaluations formulated in \citet{buathong2023bayesian}, where subfunctions are fixed and partial evaluations share consistent semantics across queries, the submodules $f_{\theta_i}$ here are themselves design variables; changing $\theta_i$ alters the mapping from input to output and latent representation space. This adds an additional layer of complexity, as intermediate outputs cannot be reused or modeled as evaluations of a fixed function.

\subsection{LLM-Driven Neural Architecture Generation}
We begin with an expert-designed initial architecture. At each subsequent iteration, we query the LLM with a concise summary of the current architecture modules and their associated performance. The LLM then produces one or a batch of new composite architecture candidates.
In this pipeline, the LLM functions as a \emph{neural architecture design agent}, generating new module configurations informed by the feedback signals from previously evaluated candidates.

For each module (including the source-specific encoders and the fusion block), the search is restricted to a module-specific architecture space typically used for that type of neural network (e.g., CNNs for image inputs, GRUs for textual instructions, transformers for time-series inputs, and FFNs for vector inputs). To illustrate this, \Cref{fig:composite_traffic} and \Cref{tab:search_space_traffic} in \Cref{apdx:experiment_setup} provide an example of a composite architecture for a mixed-autonomy traffic-control benchmark, involving transformers for time-series encoders and FFNs for vector inputs. Similarly, \Cref{fig:composite_game} and \Cref{tab:search_space_game} provide an example of a composite architecture for the MiniGrid goal-oriented task benchmark, involving an image encoder, a text encoder, and a fusion encoder, along with their corresponding search spaces. Likewise, \Cref{fig:composite_robotic} provides an example of a composite architecture for the ManiSkill robotic control benchmark, involving an RGB encoder, an information encoder, and a fusion encoder.

Finally, we convert the LLM output into executable architectures using simple tokenization and pattern matching (for implementation details, see \Cref{apdx:experiment_setup}).
For a comparison between the design of our method and other LLM-based NAS methods such as GENIUS, see \Cref{apdx:method_illustration}.

\subsection{RL Training and Evaluation}
Each generated composite architecture is trained end-to-end together with the RL agent. We train an RL algorithm (e.g., PPO) for a fixed number of interaction steps $T$. 
Compared to existing NAS methods (e.g., LLM-based method GENIUS), we exploit the intermediate outputs of composite architectures by feeding the LLM not only the task metric but also the average reward and feature information---derived from each source-specific module and reflecting its representation quality (see \Cref{apdx:method_illustration}). 
These signals provide side information for refining candidate architectures and are jointly used as performance feedback in the next iteration. 
The RL policy architecture remains fixed; only the state encoder modules vary during the search. 
This LLM--training--evaluation loop repeats until the evaluation budget is exhausted.
In the batch setting, the RL agent is trained independently for each candidate within a batch, so the overall cost scales with the number of candidates evaluated. Additional training and evaluation details are provided in \Cref{apdx:experiment_setup}.

\section{Experiment: RL-Based Mixed-Autonomy Traffic Control}
\begin{figure}[t!]
    \centering  
    \begin{subfigure}[b]{0.48\textwidth}  
        \centering  
        \includegraphics[width=\textwidth]{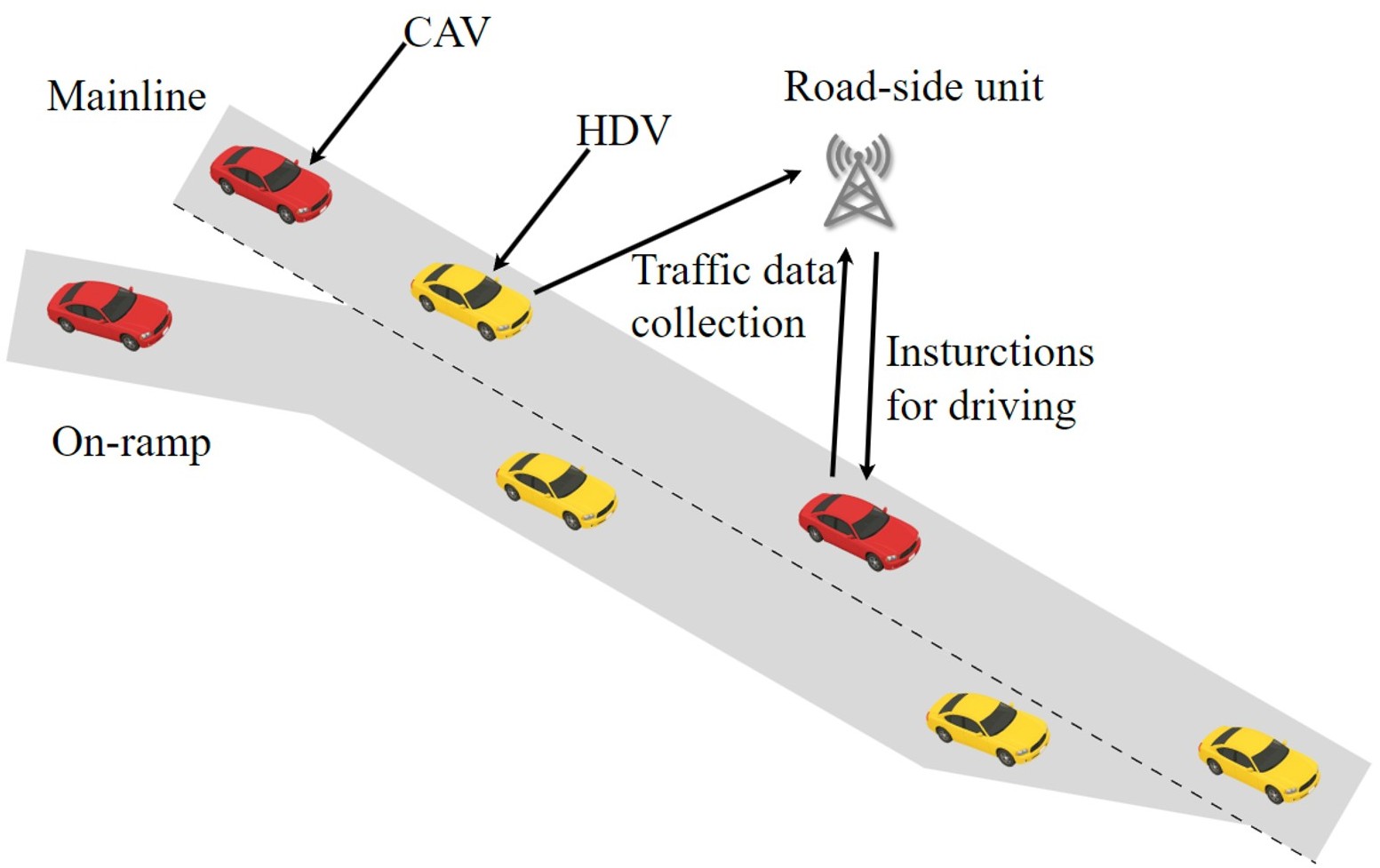}  
        \caption{RL-based mixed autonomy traffic control.} 
        \label{subfig:left_traffic}  
    \end{subfigure}
    \hfill  
    \begin{subfigure}[b]{0.48\textwidth}
        \centering
        \includegraphics[width=\textwidth]{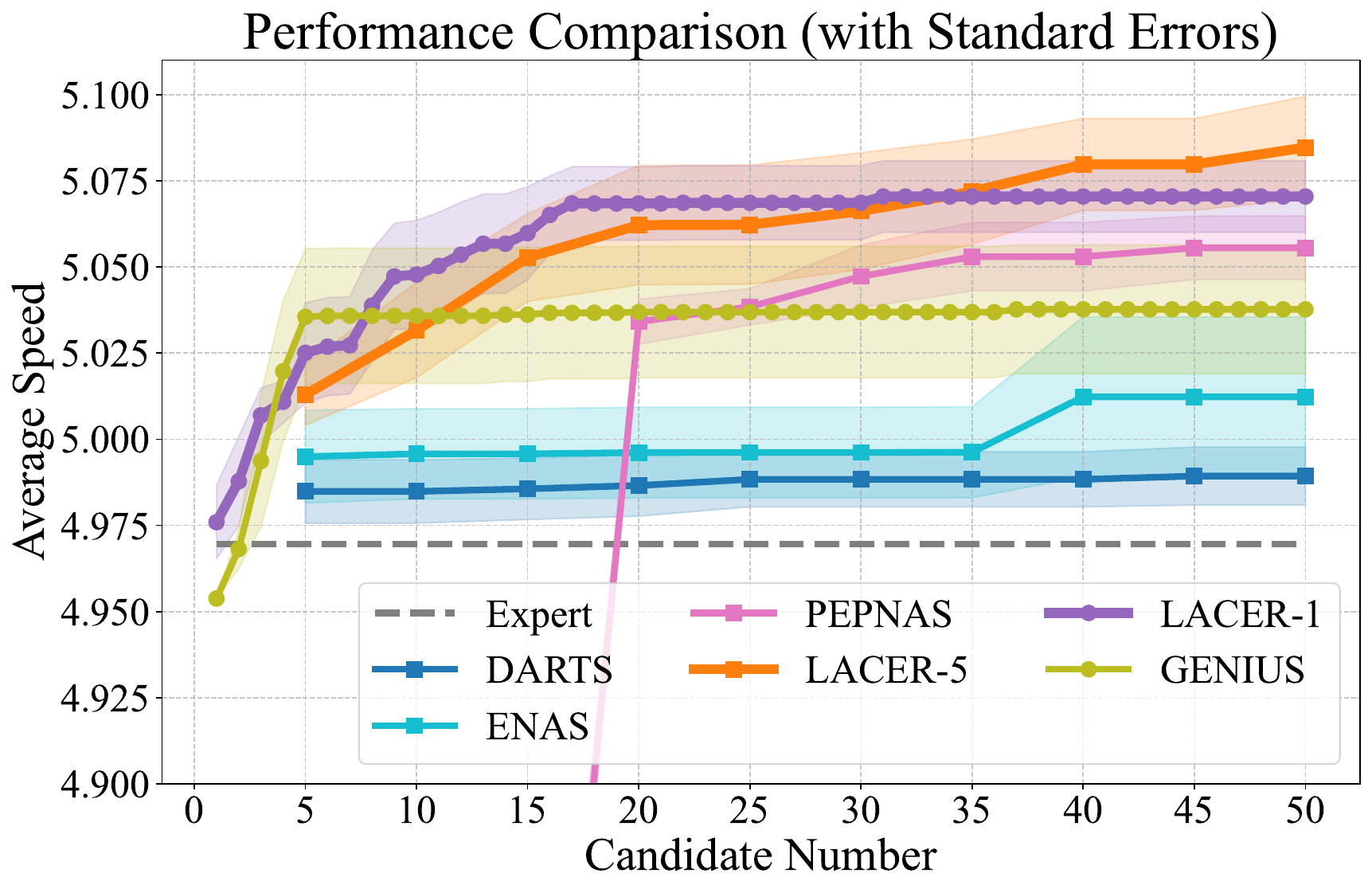}
        \caption{Performance comparison}  
        \label{subfig:right_experiment}  
    \end{subfigure}
    \caption{Left: Illustration of RL-based mixed-autonomy traffic control; Right: Comparison of performance (i.e., average traffic speed) between our two LACER variants and baselines.}
    \label{fig:combined_figures}  
\end{figure}

\paragraph{Benchmark.} We evaluate our method on an RL-based mixed-autonomy traffic control task studied in \citet{cheng2025learning} in which both connected autonomous vehicles (CAVs) and human-driven vehicles coexist in the same environment. The CAV penetration ratio is set to 0.9. At each environment step, the observation contains three distinct input sources: (i) the temporal traffic evolution of key metrics (e.g., speed, density, and flow rate), (ii) the current traffic state (lane-specific densities, speed distributions, and CAV penetration ratio), and (iii) the distribution history of the vehicle sequence. A schematic of this traffic control scenario is illustrated in \Cref{subfig:left_traffic}. The presence of \emph{multiple sources of inputs} makes this benchmark suitable for evaluating composite state encoders.

\paragraph{Baselines.} We consider three groups of baselines:
(i) \emph{Expert-designed}, i.e., encoder architectures manually specified by a domain expert;
(ii) \emph{Traditional NAS}, including DARTS~\citep{liu2018darts}, ENAS~\citep{pham2018efficient}, and PEPNAS~\citep{xue2024neural}, each generates 5 candidates per iteration;
(iii) \emph{LLM-based NAS}, including GENIUS~\citep{zheng2023can}, which uses GPT-4 to generate one candidate per iteration.

\paragraph{Evaluation metric.}
For each method, we track the \emph{average traffic speed} achieved by the \emph{best architecture evaluated so far}, and plot this task metric over the number of evaluated candidates. 
This allows us to assess the sample efficiency of the neural architecture search, i.e., how quickly each method discovers architectures with higher performance.

\paragraph{Experiment setup.} 
Following \citet{cheng2025learning}, we adopt four encoder modules: a \emph{traffic encoder}, a \emph{time encoder}, a \emph{sequence encoder}, and a \emph{fusion encoder}.
The traffic state is represented as a fixed-dimensional vector and processed by an FFN.
The remaining two inputs (temporal traffic evolution and action sequence history) are treated as time-series, and we therefore search over transformer-based architectures for their corresponding encoders.
The fusion module is also implemented as an FFN.
The architecture search space for each module is designed following the taxonomy in the survey of \citet{chitty2022neural}.
Detailed module-specific search spaces are reported in \Cref{apdx:experiment_setup}.

Each candidate architecture is trained for 200k interaction steps using PPO and then evaluated for 50k steps to obtain performance metrics, including average traffic speed, average reward, and feature information.
We consider two variants of our method: LACER-1, which generates one candidate per iteration, and LACER-5, which generates five candidates per iteration. For a fair comparison, all methods are evaluated using 50 candidates in total---10 iterations for each batch method and 50 for others.
To assess variability, each experiment is repeated with 8 random seeds.

\paragraph{Experiment results.}
\Cref{subfig:right_experiment} shows the average traffic speed as a function of the number of evaluated architecture candidates.
Both of our LACER variants significantly outperform the expert-designed architecture, traditional NAS baselines, and the LLM-based GENIUS baseline.
These results demonstrate that combining LLM-based priors with composite state encoding and richer performance signals leads to more sample-efficient architecture search in RL settings.

In \Cref{apdx:additional_results}, we also report ablation studies to analyze the effect of different design choices in our pipeline,
including:
(i) additionally providing the task metric (e.g., average traffic speed) of the initial expert-designed architecture;
(ii) using only the task metric versus also including the average reward in the feedback;
(iii) additionally providing the feature information (representation quality).

\section{Conclusion and Future Directions}
In this work, we proposed LACER, an LLM-driven composite NAS pipeline that automatically discovers effective state encoders for multi-source RL, achieving better task performance than traditional and LLM-based NAS baselines. 
In future work, we plan to apply LACER in broader applications such as goal-oriented tasks and robotic control with visual, textual and sensor inputs. 

\section{Acknowledgments and Funding Disclosure}
This work was in part supported by the National Natural Science Foundation of China under Grant 62473250.
The authors acknowledge the support of undergraduate researchers Junping Li, Aghamatlab Akbarzade, Yuchen Jiang for their contributions to preliminary experimentation.

\bibliographystyle{plainnat}
\bibliography{references}

@article{liu2018darts,
  title={Darts: Differentiable architecture search},
  author={Liu, Hanxiao and Simonyan, Karen and Yang, Yiming},
  journal={arXiv preprint arXiv:1806.09055},
  year={2018}
}

@inproceedings{pham2018efficient,
  title={Efficient neural architecture search via parameters sharing},
  author={Pham, Hieu and Guan, Melody and Zoph, Barret and Le, Quoc and Dean, Jeff},
  booktitle={International conference on machine learning},
  pages={4095--4104},
  year={2018},
  organization={PMLR}
}

@article{xue2024neural,
  title={Neural architecture search with progressive evaluation and sub-population preservation},
  author={Xue, Yu and Zha, Jiajie and Pelusi, Danilo and Chen, Peng and Luo, Tao and Zhen, Liangli and Wang, Yan and Wahib, Mohamed},
  journal={IEEE Transactions on Evolutionary Computation},
  year={2024},
  publisher={IEEE}
}

@article{chitty2022neural,
  title={Neural architecture search for transformers: A survey},
  author={Chitty-Venkata, Krishna Teja and Emani, Murali and Vishwanath, Venkatram and Somani, Arun K},
  journal={IEEE Access},
  volume={10},
  pages={108374--108412},
  year={2022},
  publisher={IEEE}
}

@article{zheng2023can,
  title={Can gpt-4 perform neural architecture search?},
  author={Zheng, Mingkai and Su, Xiu and You, Shan and Wang, Fei and Qian, Chen and Xu, Chang and Albanie, Samuel},
  journal={arXiv preprint arXiv:2304.10970},
  year={2023}
}

@inproceedings{nasir2024llmatic,
  title={Llmatic: neural architecture search via large language models and quality diversity optimization},
  author={Nasir, Muhammad Umair and Earle, Sam and Togelius, Julian and James, Steven and Cleghorn, Christopher},
  booktitle={proceedings of the Genetic and Evolutionary Computation Conference},
  pages={1110--1118},
  year={2024}
}

@inproceedings{zhou2025design,
  title={Design principle transfer in neural architecture search via large language models},
  author={Zhou, Xun and Wu, Xingyu and Feng, Liang and Lu, Zhichao and Tan, Kay Chen},
  booktitle={Proceedings of the AAAI Conference on Artificial Intelligence},
  volume={39},
  number={21},
  pages={23000--23008},
  year={2025}
}

@article{cai2025seki,
  title={SEKI: Self-Evolution and Knowledge Inspiration based Neural Architecture Search via Large Language Models},
  author={Cai, Zicheng and Tang, Yaohua and Lai, Yutao and Wang, Hua and Chen, Zhi and Chen, Hao},
  journal={arXiv preprint arXiv:2502.20422},
  year={2025}
}

@article{yan2024efficient,
  title={Efficient reinforcement learning with large language model priors},
  author={Yan, Xue and Song, Yan and Feng, Xidong and Yang, Mengyue and Zhang, Haifeng and Ammar, Haitham Bou and Wang, Jun},
  journal={arXiv preprint arXiv:2410.07927},
  year={2024}
}

@article{schoepp2025evolving,
  title={The evolving landscape of llm-and vlm-integrated reinforcement learning},
  author={Schoepp, Sheila and Jafaripour, Masoud and Cao, Yingyue and Yang, Tianpei and Abdollahi, Fatemeh and Golestan, Shadan and Sufiyan, Zahin and Zaiane, Osmar R and Taylor, Matthew E},
  journal={arXiv preprint arXiv:2502.15214},
  year={2025}
}

@inproceedings{falkner2018bohb,
  title={BOHB: Robust and efficient hyperparameter optimization at scale},
  author={Falkner, Stefan and Klein, Aaron and Hutter, Frank},
  booktitle={International conference on machine learning},
  pages={1437--1446},
  year={2018},
  organization={PMLR}
}

@inproceedings{white2021bananas,
  title={Bananas: Bayesian optimization with neural architectures for neural architecture search},
  author={White, Colin and Neiswanger, Willie and Savani, Yash},
  booktitle={Proceedings of the AAAI conference on artificial intelligence},
  volume={35},
  number={12},
  pages={10293--10301},
  year={2021}
}

@article{cao2024survey,
  title={Survey on large language model-enhanced reinforcement learning: Concept, taxonomy, and methods},
  author={Cao, Yuji and Zhao, Huan and Cheng, Yuheng and Shu, Ting and Chen, Yue and Liu, Guolong and Liang, Gaoqi and Zhao, Junhua and Yan, Jinyue and Li, Yun},
  journal={IEEE Transactions on Neural Networks and Learning Systems},
  year={2024},
  publisher={IEEE}
}

@inproceedings{cheng2025learning,
  author    = {X. Cheng and L. Jin},
  title     = {Learning-Based Vehicle Sequencing for On-ramp Merging in Mixed Traffic},
  booktitle = {Proceedings of the 23rd IEEE International Conference on Industrial Informatics (INDIN)},
  year      = {2025},
  publisher = {IEEE},
  pages     = {0--0}
}

@article{chevalier2023minigrid,
  title={Minigrid \& miniworld: Modular \& customizable reinforcement learning environments for goal-oriented tasks},
  author={Chevalier-Boisvert, Maxime and Dai, Bolun and Towers, Mark and Perez-Vicente, Rodrigo and Willems, Lucas and Lahlou, Salem and Pal, Suman and Castro, Pablo Samuel and Terry, Jordan},
  journal={Advances in Neural Information Processing Systems},
  volume={36},
  pages={73383--73394},
  year={2023}
}

@article{balandat2020botorch,
  title={BoTorch: A framework for efficient Monte-Carlo Bayesian optimization},
  author={Balandat, Maximilian and Karrer, Brian and Jiang, Daniel and Daulton, Samuel and Letham, Ben and Wilson, Andrew G and Bakshy, Eytan},
  journal={Advances in neural information processing systems},
  volume={33},
  pages={21524--21538},
  year={2020}
}

@article{tao2024maniskill3,
  title={Maniskill3: Gpu parallelized robotics simulation and rendering for generalizable embodied ai},
  author={Tao, Stone and Xiang, Fanbo and Shukla, Arth and Qin, Yuzhe and Hinrichsen, Xander and Yuan, Xiaodi and Bao, Chen and Lin, Xinsong and Liu, Yulin and Chan, Tse-kai and others},
  journal={arXiv preprint arXiv:2410.00425},
  year={2024}
}

@article{wang2024llm,
  title={LLM-empowered state representation for reinforcement learning},
  author={Wang, Boyuan and Qu, Yun and Jiang, Yuhang and Shao, Jianzhun and Liu, Chang and Yang, Wenming and Ji, Xiangyang},
  journal={arXiv preprint arXiv:2407.13237},
  year={2024}
}

@article{buathong2023bayesian,
  title={Bayesian optimization of function networks with partial evaluations},
  author={Buathong, Poompol and Wan, Jiayue and Astudillo, Raul and Daulton, Samuel and Balandat, Maximilian and Frazier, Peter I},
  journal={arXiv preprint arXiv:2311.02146},
  year={2023}
}

\newpage
\appendix
\section{Methodology Illustration}
\label{apdx:method_illustration}

\paragraph{Paradigm comparison.}

\begin{figure}[htp]
    \centering
    \includegraphics[width=10cm]{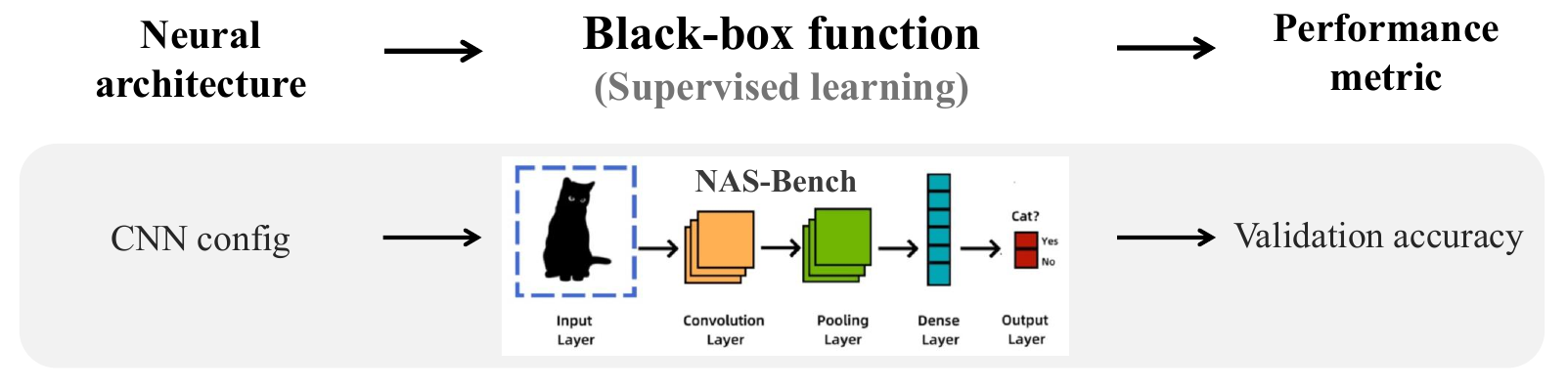} 
    \caption{Paradigm of traditional NAS for supervised learning.}
    \label{fig:tnas}
\end{figure}

Unlike traditional NAS methods for supervised learning (\Cref{fig:tnas}), our approach is designed to optimize multi-source state encoders in RL. 

\begin{figure}[htp]
    \centering
    \includegraphics[width=11cm]{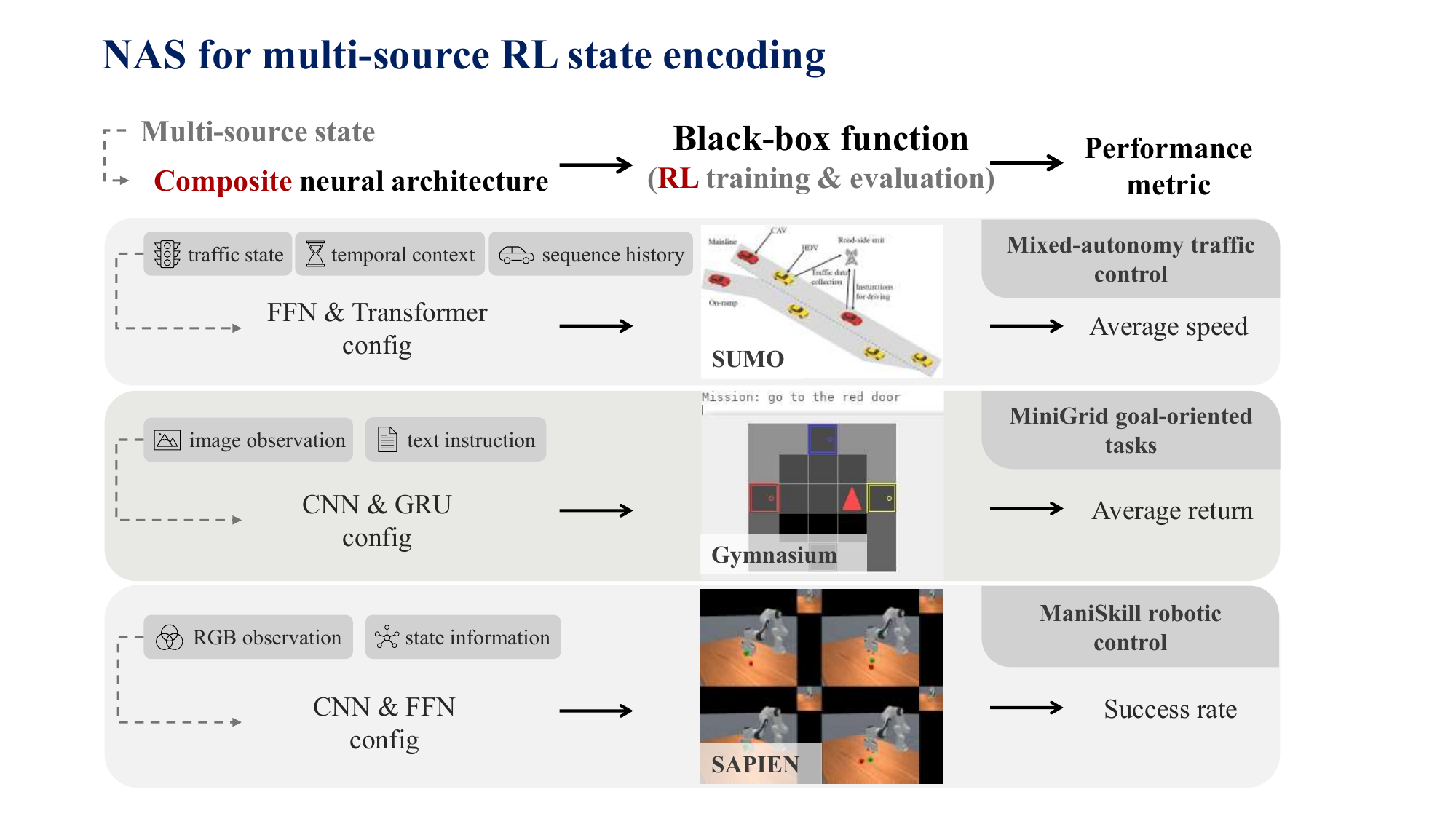} 
    \caption{Paradigm of composite NAS for multi-source RL state encoding (ours).}
    \label{fig:cnas}
\end{figure}

This leads to composite NAS with composite architecture design of the multi-source RL state encoder and corresponding search space (\Cref{fig:cnas}).

\paragraph{Pipeline comparison.}

\begin{figure}[htp]
    \centering
    \includegraphics[width=10cm]{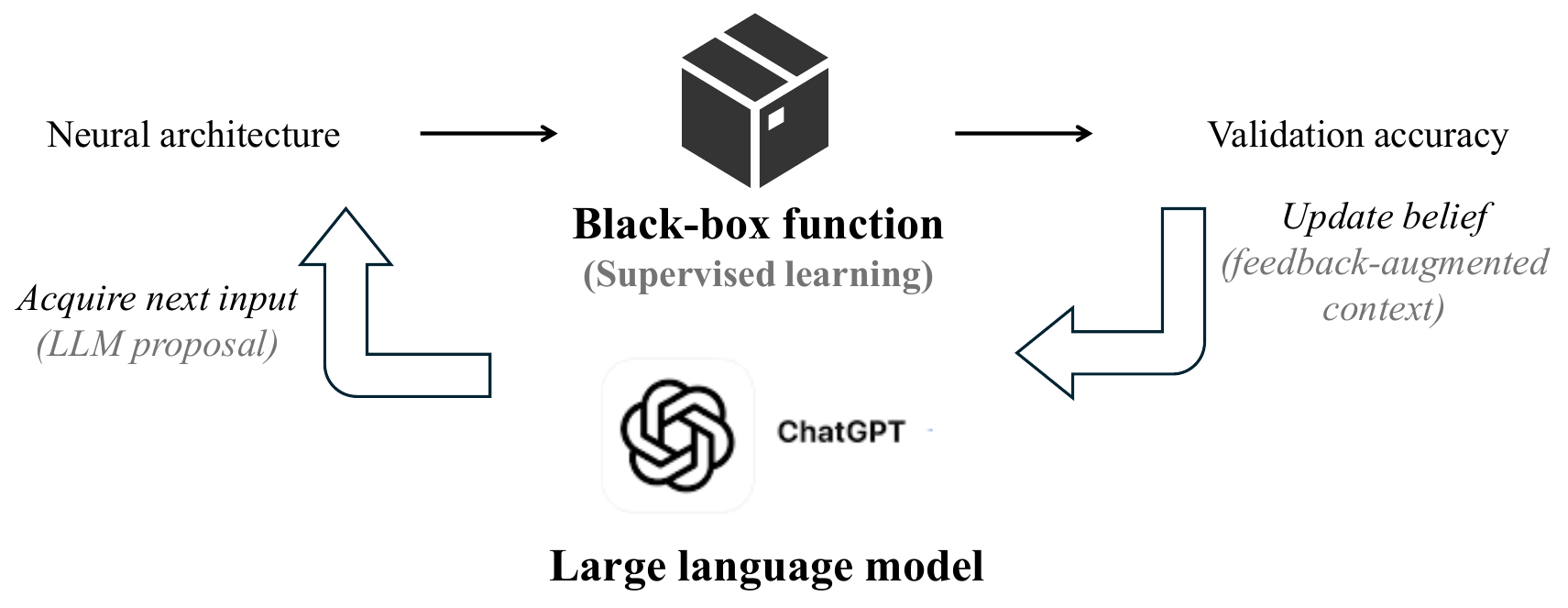} 
    \caption{Pipeline of GENIUS, which only uses performance metric as update belief for supervised learning.}
    \label{fig:genius_loop}
\end{figure}

Compared to other LLM-based NAS methods (e.g., GENIUS in \Cref{fig:genius_loop}), our approach is designed to enhance both sample efficiency and solution quality when searching for composite neural architectures for state encoders in RL. 

\begin{figure}[htp]
    \centering
    \includegraphics[width=11cm]{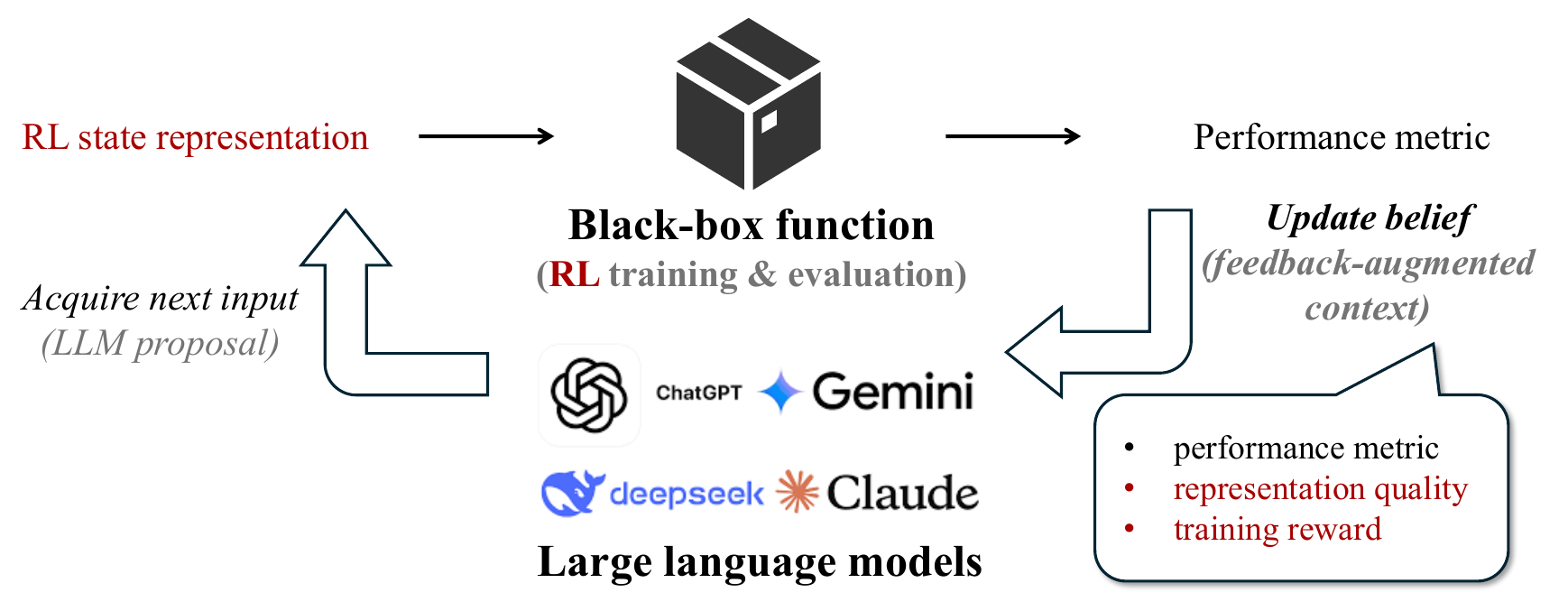} 
    \caption{Pipeline of LACER (our method), which exploits side information on source-specific encoders beyond performance metric for RL.}
    \label{fig:lacer_loop}
\end{figure}

To achieve this, we incorporate evaluation metrics of the initial architecture into the initial prompt to provide richer prior context. Additionally, beyond standard task metric, we introduce two supplementary performance signals---\emph{average reward} and \emph{feature information} (i.e., representation quality)---as comprehensive feedback to the LLM, enabling iterative refinement of candidate architectures (\Cref{fig:lacer_loop}).

\paragraph{Performance signals.}

Our method’s performance signals include three components: (i) task metric: average speed in the mixed-autonomy traffic control setting, served as the ultimate indicator of target task; (ii) average reward: incorporated as feedback given that the candidate architectures are for RL training, which characterizes key RL properties like convergence efficiency; (iii) feature information: quantified via mutual information (defined as $I(X;Y)=H(X)-H(X \mid Y)$ for random variables $X$ and $Y$) and redundancy (defined as $R(X;Y)=H(X)+H(Y)-H(X,Y)$), served as a direct measure of representation quality for the composite state representation architecture. Specifically, we compute mutual information for feature pairs: time data before/after time encoder, traffic data before/after traffic encoder, sequence data before/after sequence encoder, and encoded features from time/traffic/sequence encoders with fused features after fusion encoder to comprehensively address the composite nature of the state encoder with multiple modules.

\paragraph{Prompt construction.}

The iterative prompt construction process used in our method is detailed in \Cref{alg:prompt_construction}. The conversation history $\mathcal{H}$ is strategically pruned to retain only essential information---including initial architecture, performance signals, task description, and search space definition---while eliminating redundant and non-structural content.
This reduces noise and mitigates potential LLM forgetfulness in long interactions. 
In the prompt, different roles are explicitly distinguished: the 
\emph{assistant} role logs the LLM’s responses in $\mathcal{H}$, while the 
\emph{system} and \emph{user} roles provide setup and queries, respectively.
The initial user prompt $\mathcal{U}_0$ is designed to activate the LLM's prior knowledge through task description and structured search space modeling, while iterative prompts $\mathcal{U}_i$ reinforce the search space and constraints to maintain robustness throughout generations.

\begin{algorithm}
\caption{Prompt Construction of LACER}
\label{alg:prompt_construction}
\begin{algorithmic}[1]
\REQUIRE
System prompt $\mathcal{S}$,
Task description $\mathcal{D}$,
Search space $\mathcal{X}$,
Request $\mathcal{R}$,
Initial architecture $a_0$ and its performance $p_0$,
Max iterations $N$
\ENSURE Candidate architecture lists $\mathcal{L}_1, \dots, \mathcal{L}_N$
\STATE Initialize conversation history $\mathcal{H} \gets \emptyset$
\STATE $\mathcal{U}_0 \gets \mathcal{D} + \mathcal{X} + a_0 + p_0 + \mathcal{R}$ \COMMENT{First iteration user prompt}
\STATE $\text{Prompt}_0 \gets \mathcal{S} + \mathcal{U}_0$
\STATE $\mathcal{L}_1 \gets \text{LLM}(\text{Prompt}_0)$
\STATE $\vec{v}_{\text{raw}} \gets \text{ParseLLMResponse}(\mathcal{L}_1)$ \COMMENT{Using \Cref{alg:parse_llm}}
\STATE Append ${\textit{system}: \mathcal{S}, \textit{user}: \mathcal{U}_0, \textit{assistant}: \mathcal{L}_1}$ to $\mathcal{H}$
\FOR{$i = 1$ to $N-1$}
\STATE $\mathcal{U}_i \gets \mathcal{P}_{i-1} + \mathcal{X} + \mathcal{R}$ \COMMENT{Subsequent user prompts}
\STATE $\text{Prompt}\ i \gets \mathcal{H} + \mathcal{U}_i$
\STATE $\mathcal{L}_{i+1} \gets \text{LLM}(\text{Prompt}_i)$
\STATE $\vec{v}_{\text{raw}} \gets \text{ParseLLMResponse}(\mathcal{L}_{i+1})$ \COMMENT{Using \Cref{alg:parse_llm}}
\STATE Append ${\textit{user}: \mathcal{U}_i, \textit{assistant}: \mathcal{L}_{i+1}}$ to $\mathcal{H}$
\STATE Train and evaluate each architecture in $\mathcal{L}_{i+1}$ in RL framework
\ENDFOR
\end{algorithmic}
\end{algorithm}

\section{Experimental Setup and Implementation Details}
\label{apdx:experiment_setup}

\subsection{RL-based mixed-autonomy traffic control}
All experiments were run on the NYU Greene high-performance computing cluster, using a single NVIDIA GPU per run (either a Quadro RTX~8000 with 48\,GB of memory or a Tesla V100 with 32\,GB of memory), together with 16 CPU cores and 32\,GB of RAM. Unless otherwise specified, each configuration was repeated with 8 random seeds, and we report the mean performance with error bars corresponding to two times the standard error across seeds.


\paragraph{Composite architecture design of the RL state encoder.}

\begin{figure}[htp]
    \centering
    \includegraphics[width=12cm]{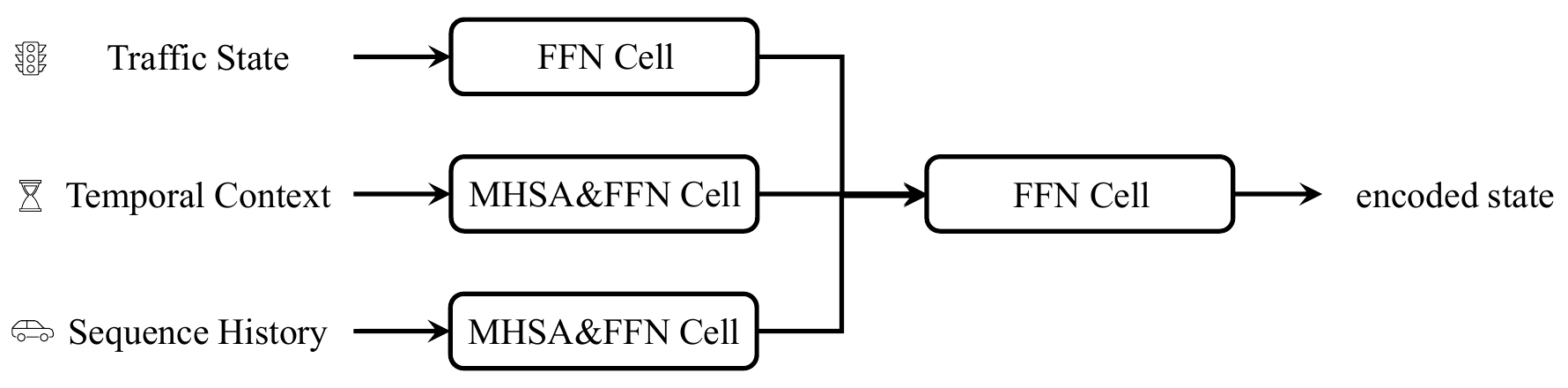} 
    \caption{Composite architecture design of the RL state encoder for mixed-autonomy traffic control.}
    \label{fig:composite_traffic}
\end{figure}

Following \citet{cheng2025learning}, the architecture of the state encoder includes a time encoder, traffic encoder, and sequence encoder for respective data encoding while the fusion encoder processes their concatenated outputs to generate the encoded state for RL training, as illustrated in \Cref{fig:composite_traffic}.  All modules use Transformer (chosen over recurrent alternatives such as LSTMs, given its superior performance on sequential data) with FFNs. Time and sequence encoders add multi-head self-attention (MHSA) to capture their higher complexity, temporal variability, and dynamics, unlike preprocessed, macro-level, weakly temporal traffic data, and the fusion encoder which focused on integration without extra temporal processing uses only FFN.

\paragraph{Module-specific search spaces.}

Based on the state encoder architecture with multiple modules of distinct functions and specific architecture types in \Cref{fig:composite_traffic}, we define module-specific search spaces, as detailed in \Cref{tab:module_parameters_traffic}. Following \citet{chitty2022neural}, key design choices commonly used in neural architecture search for Transformers are included: hidden layer dimension (denoted as "dimension" in the table), dimension expansion ratio ("ratio"), and number of neural network layers ("depth") for the FFN of each module. For modules with MHSA, the number of attention heads ("heads") is additionally included; for modules with only FFN, the type of activation function ("activation") is included instead. Common value ranges are set for all design choices, resulting in a total of approximately 26 million possible architectures throughout the search space.

\begin{table}[htbp]
    \centering
    \caption{Module-specific search spaces where bold values denote the configurations used by the Expert baseline.}
    \label{tab:module_parameters_traffic}
    \begin{tabularx}{\linewidth}{
        >{\hsize=0.7\hsize}X  
        >{\hsize=1.0\hsize}X  
        >{\hsize=1.5\hsize}X  
        >{\hsize=1.0\hsize}X  
        >{\hsize=0.6\hsize}X  
        >{\hsize=0.6\hsize}X  
    }
        \toprule
        \textbf{Module} & \textbf{Operation} & \textbf{Heads / Activation} & \textbf{Dimension} & \textbf{Ratio} & \textbf{Depth} \\
        \midrule
        Time & MHSA, FFN & $\{\textbf{2}, 4, 8\}$ & $\{\textbf{8}, 16, 32\}$ & $\{1, \textbf{2}, 4\}$ & $\{1, \textbf{2}, 3\}$ \\
        Traffic & FFN & $\{\text{\textbf{relu}}, \text{gelu}, \text{swish}\}$ & $\{16, \textbf{32}, 64\}$ & $\{\textbf{1}, 2, 4\}$ & $\{\textbf{1}, 2, 3\}$ \\
        Sequence & MHSA, FFN & $\{2, \textbf{4}, 8\}$ & $\{8, \textbf{16}, 32\}$ & $\{1, \textbf{2}, 4\}$ & $\{1, \textbf{2}, 3\}$ \\
        Fusion & FFN & $\{\text{\textbf{relu}}, \text{gelu}, \text{swish}\}$ & $\{64, \textbf{128}, 256\}$ & $\{\textbf{1}, 2, 4\}$ & $\{\textbf{1}, 2, 3\}$ \\
        \bottomrule
    \end{tabularx}
\label{tab:search_space_traffic}
\end{table}

\paragraph{Baseline alignment.}
\label{apdx:baseline_alignment}

Traditional NAS methods which we consider as baselines such as DARTS, PEPNAS, and ENAS were originally designed for computer vision tasks like image classification, where performance is measured by accuracy and the concept of sample size is applicable. However, in RL scenarios, accuracy is absent, and the concept of sample size differs. Thus, when applying these methods to neural architecture search for state encoders in RL, certain corresponding mappings are required: (i) Performance: The accuracy used in ENAS and PEPNAS is replaced here by average speed; similarly, the gradient in DARTS, which reflects validation performance, is also mapped to average speed. (ii) Sample size: In PEPNAS, the sample size for validating candidates within each generation increases incrementally, which corresponds here to an incremental increase in training steps when validating candidates within each generation.

\paragraph{RL training and evaluation details.}

\begin{figure}[htbp]
    \centering
    \begin{subfigure}[b]{0.49\textwidth}
        \centering
        \includegraphics[width=\textwidth]{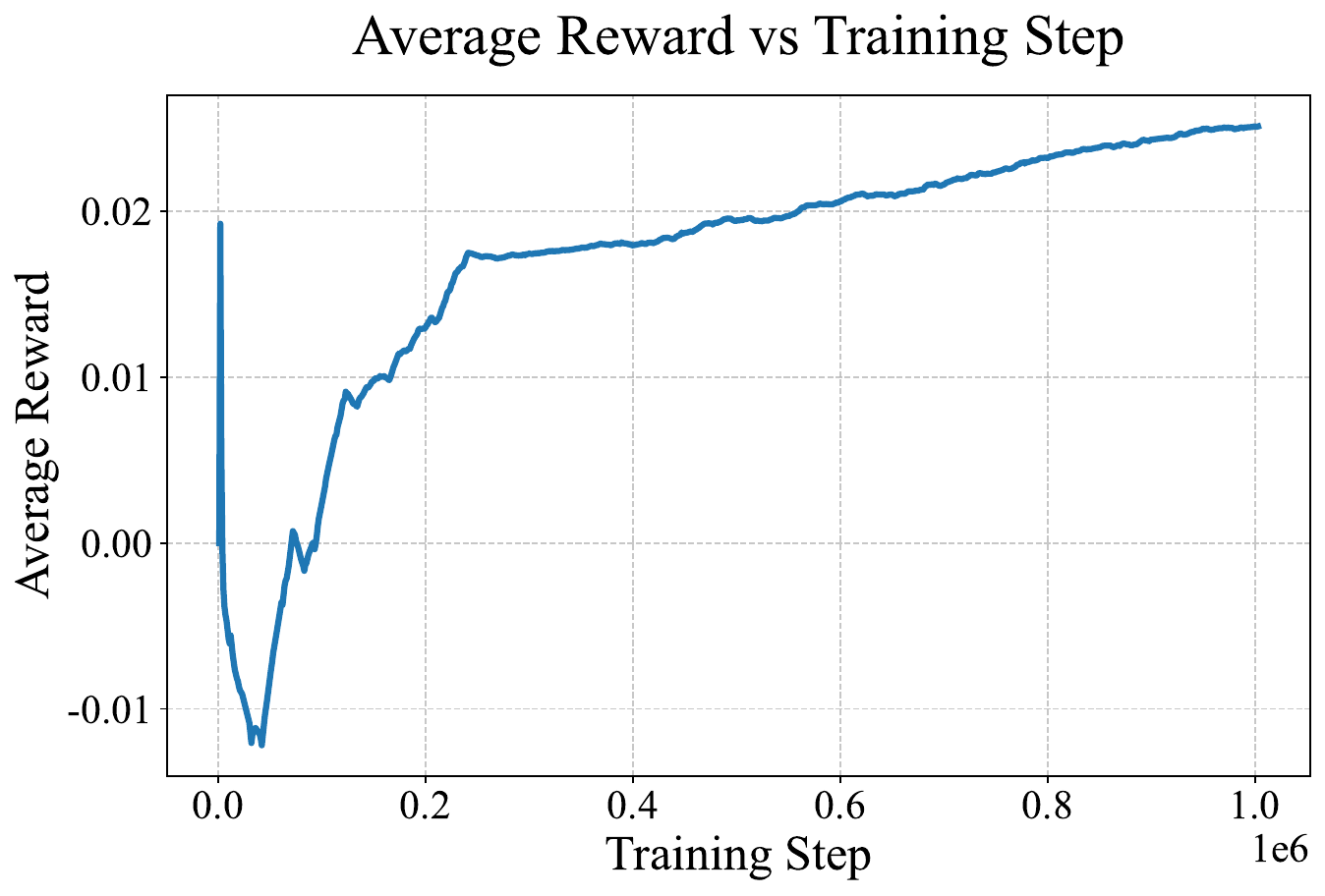} 
        \caption{Average reward over 1M steps}
        \label{fig:reward_1M}
    \end{subfigure}
    \hfill
        \begin{subfigure}[b]{0.49\textwidth}
        \centering
        \includegraphics[width=\textwidth]{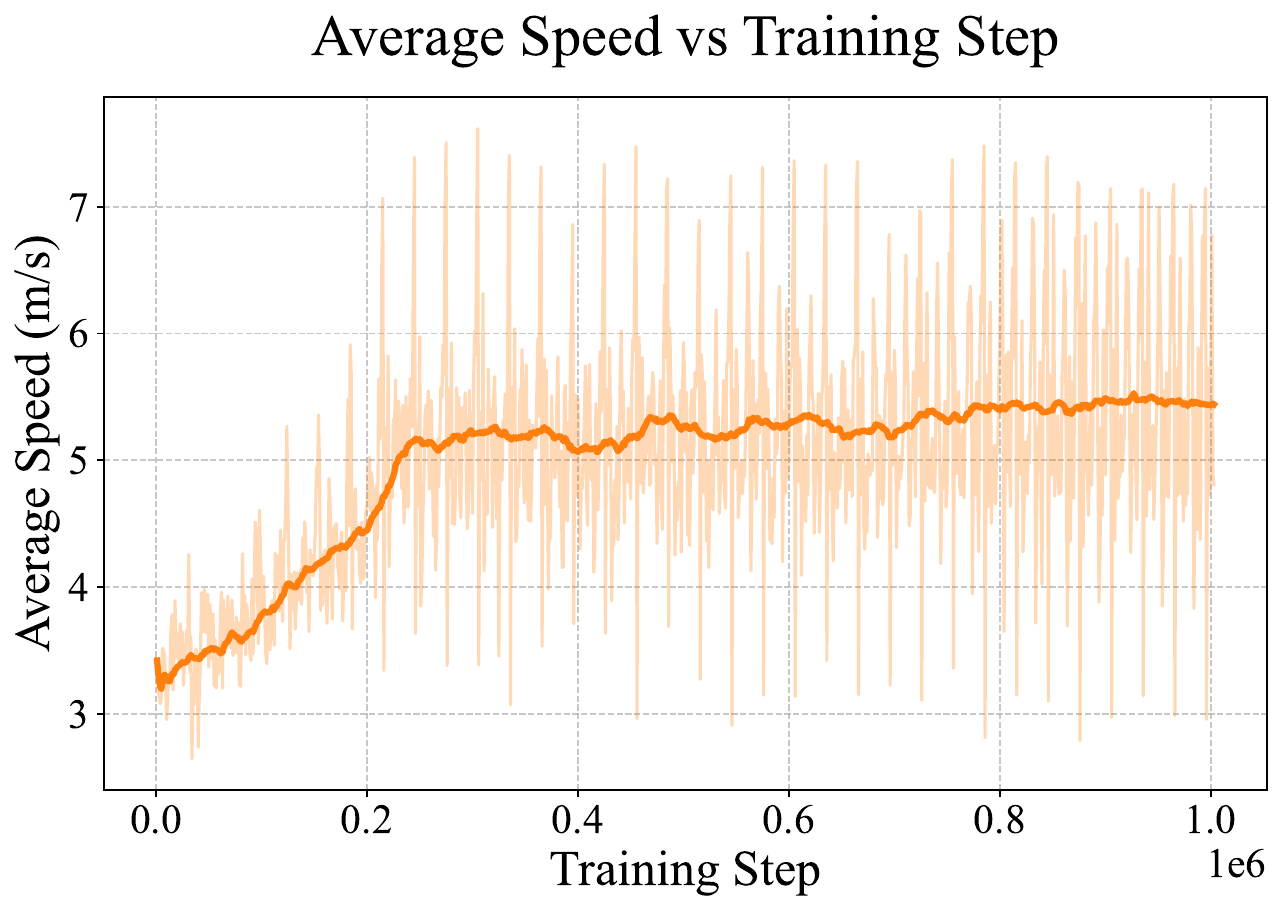} 
        \caption{Average speed over 1M steps}
        \label{fig:speed_1M}
    \end{subfigure}
    \hfill
    \begin{subfigure}[b]{0.49\textwidth}
        \centering
        \includegraphics[width=\textwidth]{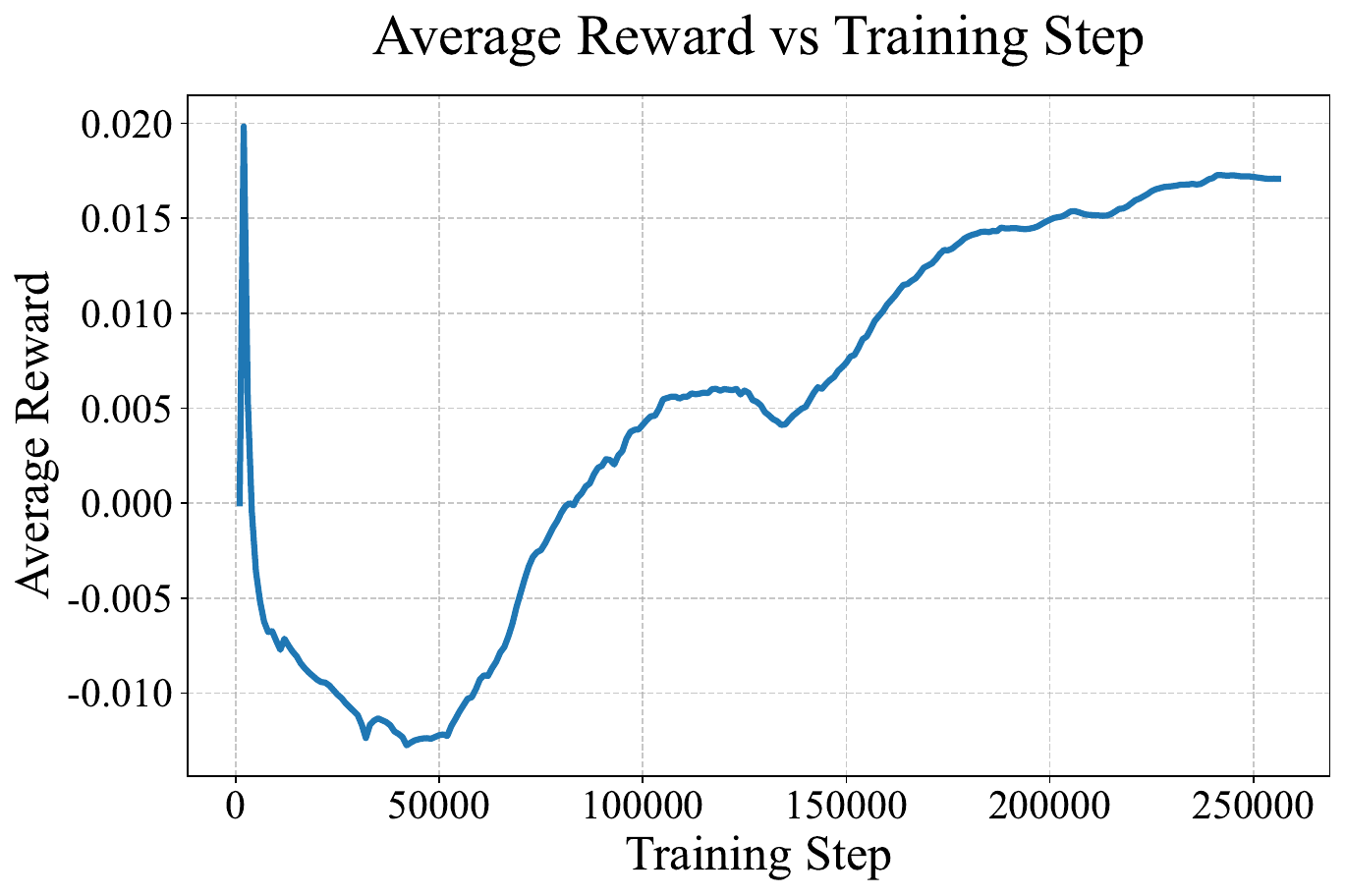} 
        \caption{Average reward over 250,000 steps}
        \label{fig:reward_250k}
    \end{subfigure}
    \begin{subfigure}[b]{0.49\textwidth}
        \centering
        \includegraphics[width=\textwidth]{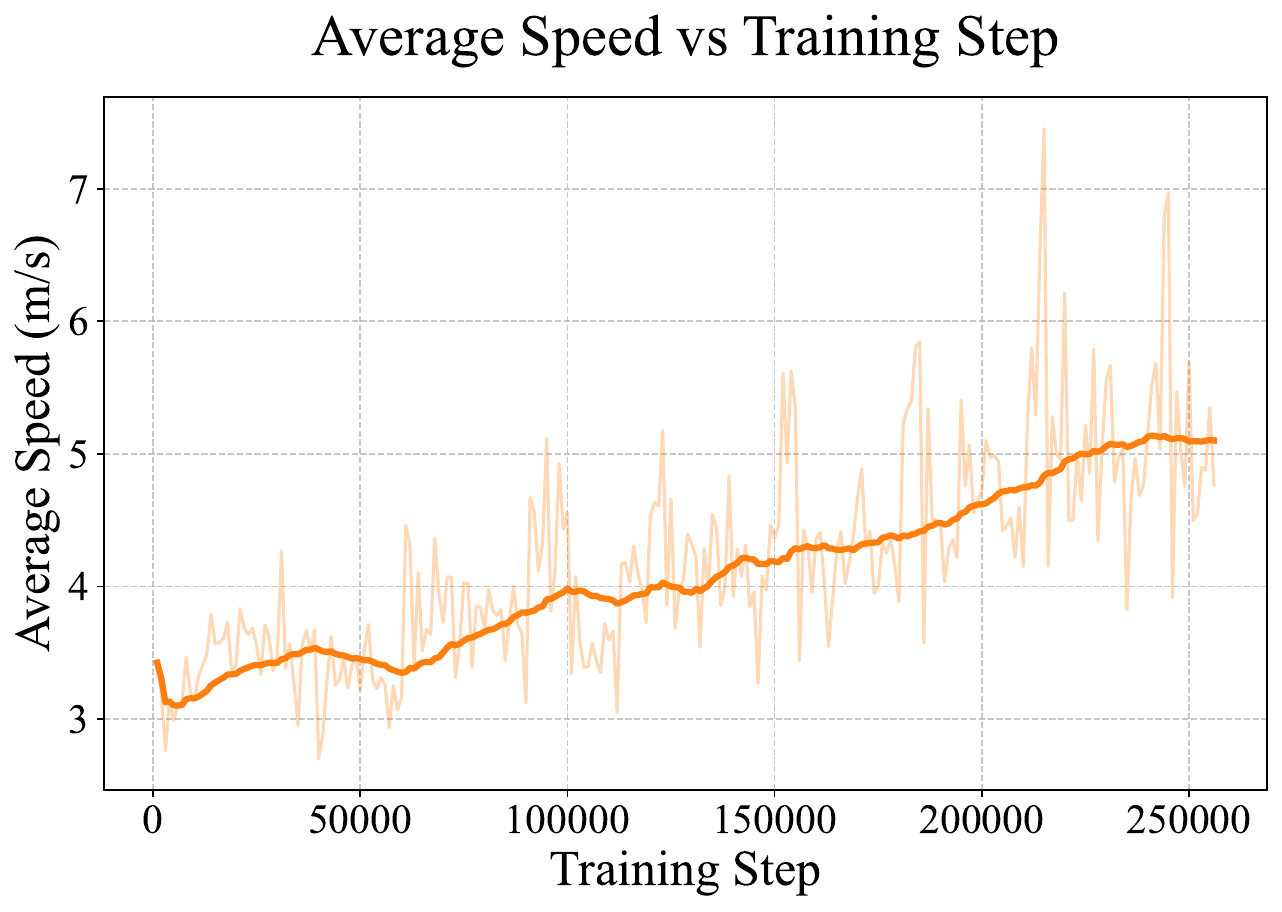} 
        \caption{Average speed over 250,000 steps}
        \label{fig:speed_250k}
    \end{subfigure}
    \caption{Average reward and average speed during RL training.}
    \label{fig:rl_training_analysis}
\end{figure}

RL training requires sufficient steps for policy convergence, typically manifested by reward. In the mixed-autonomy traffic control settings simulated via SUMO, traffic flow arrives with a fixed periodic distribution, causing observed average vehicle speed to exhibit corresponding periodicity. Thus, RL policy evaluation also requires adequate steps to encompass multiple such cycles. To balance RL training convergence, evaluation comprehensiveness, and training cost, we analyzed the average reward and average speed (recorded every 1k steps over 1M steps of baseline RL training) as shown in \Cref{fig:rl_training_analysis}. The results indicate that the average reward converges around 200k steps, while the average speed exhibits a periodicity of approximately 25k steps. Based on this, each candidate state encoder architecture was evaluated by integrating it into the RL framework for 200k steps of training and 50k steps of evaluation.

\paragraph{Parsing of LLM response.}

To ensure automation of the LLM-based neural architecture search process, we adopt an algorithm to parse the structured natural language output from LLM into a vector of design choices, as presented in \Cref{alg:parse_llm}. Firstly, the LLM is instructed to frame its architectural descriptions using a specific prefix (e.g., 'New Architecture'), which allows for the reliable extraction of the relevant text segment from its complete response. This segment is subsequently tokenized and parsed using a set of regular expression patterns that map directly to the design choices of the search space (e.g., heads, depth). The algorithm outputs the raw parsed values, which are then used directly to instantiate the state encoder for reinforcement learning.

\begin{algorithm}
\caption{Parse LLM Response to Architectures}
\label{alg:parse_llm}
\begin{algorithmic}[1]
\REQUIRE LLM response $R$, Prefix string $P$, Pattern set $\mathcal{P}$ (regex patterns for each parameter)
\ENSURE Raw parameter vector $\vec{v}_{\text{raw}}$
\STATE $\text{text\_block} \gets \text{ExtractTextAfterPrefix}(R, P)$ \COMMENT{Get the structured output}
\STATE $\text{tokens} \gets \text{Tokenize}(\text{text\_block})$ \COMMENT{Break into processable units}
\STATE $\vec{v}_{\text{raw}} \gets [\ ]$ \COMMENT{Initialize an empty list for design choices}
\FOR{{\bf each} pattern $p_i \in \mathcal{P}$}
    \STATE $value \gets \text{ApplyRegex}(p_i, \text{tokens})$ \COMMENT{Match pattern against tokens}
    \STATE $\vec{v}_{\text{raw}}.append(value)$ \COMMENT{Append the parsed value}
\ENDFOR
\RETURN $\vec{v}_{\text{raw}}$
\end{algorithmic}
\end{algorithm}

\paragraph{LLM and temperature parameter selection.}

\begin{figure}[htp]
    \centering
    \includegraphics[width=0.7\textwidth]{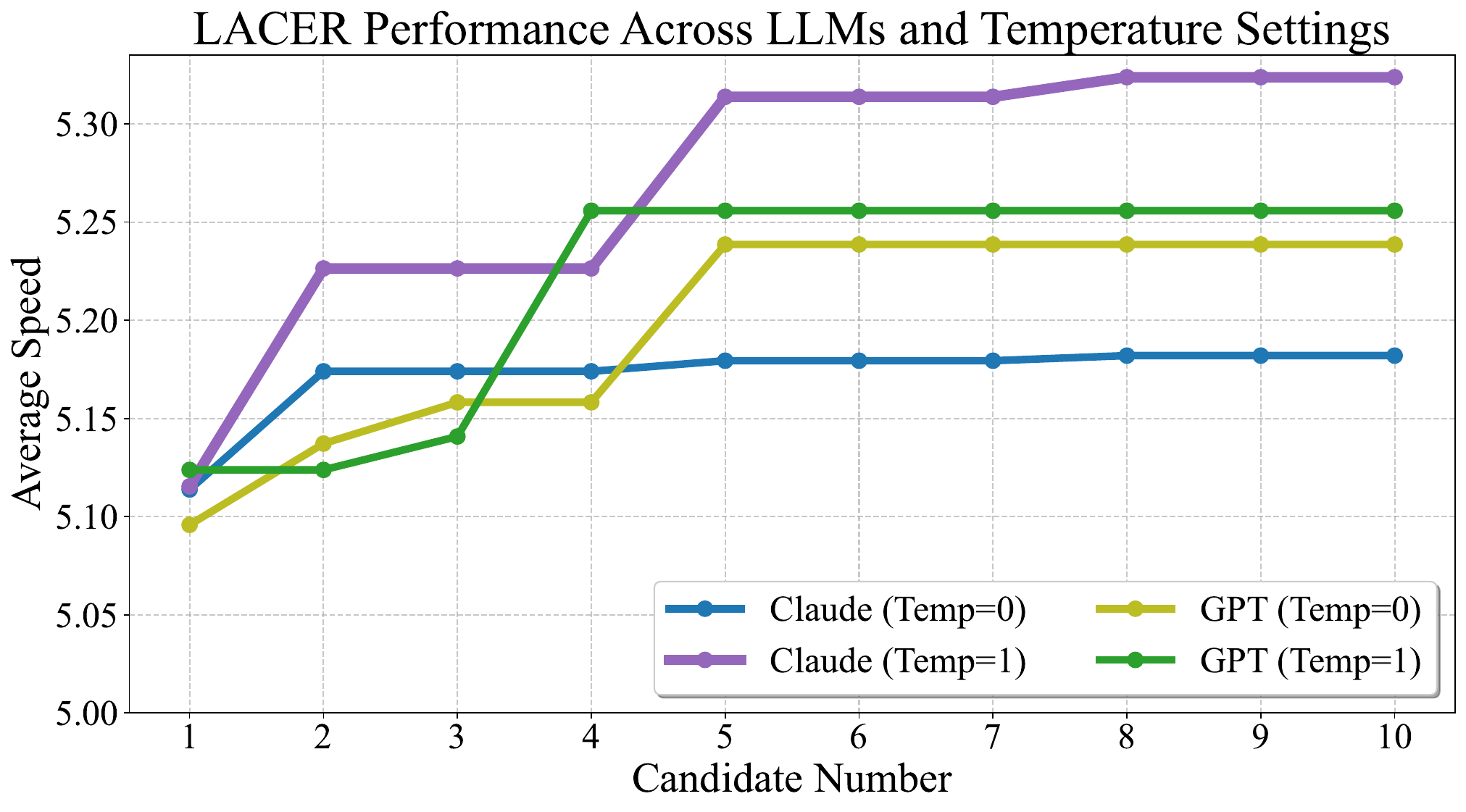} 
    \caption{Performance of LACER-1 under different LLM models and temperature settings.}
    \label{fig:type_temperature}
\end{figure}

For the proposed LACER method (applied to NAS for RL state encoders), LLM type and temperature (a hyperparameter regulating LLM output randomness: higher values enhance diversity, lower values improve determinism) significantly affect task performance. To analyze the impact of design choices, such as the base LLM and its temperature, experiments were conducted on LACER using two representative LLMs (Claude Sonnet 4.0, GPT-4) under temperature configurations of 0.0 and 1.0. As shown in \Cref{fig:type_temperature}, LACER achieved optimal performance with Claude Sonnet 4.0 (temperature = 1.0), which was thus adopted in the main experiments.

\subsection{MiniGrid goal-oriented tasks}

\paragraph{Architecture of the state encoder.}

\begin{figure}[htp]
    \centering
    \includegraphics[width=12cm]{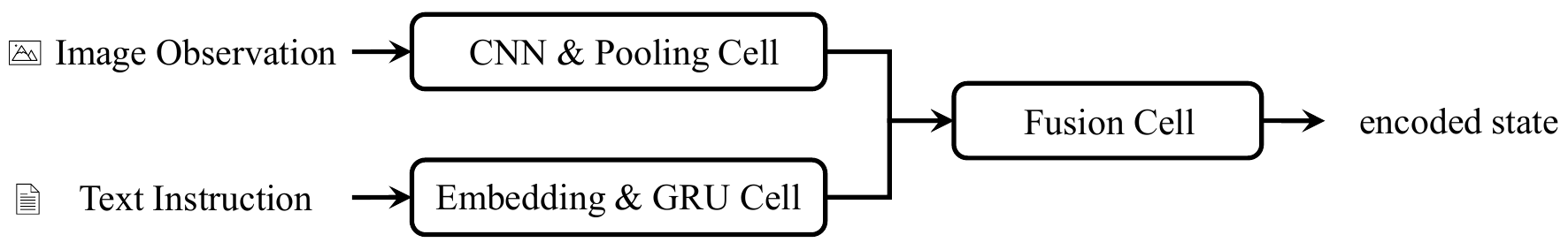} 
    \caption{Composite architecture of the state encoder in the MiniGrid goal-oriented tasks}
    \label{fig:composite_game}
\end{figure}

Following \citet{chevalier2023minigrid}, the architecture of the state encoder includes an image encoder and a text encoder for respective data encoding while the fusion encoder processes their concatenated outputs to generate the encoded state for RL training, as illustrated in \Cref{fig:composite_game}.  The image encoder uses convolution neural networks (CNNs) added with a pooling layer to process image observation. The text encoder uses an embedding layer added with a gate recurrent unit (GRU) to represent the text instruction. The fusion encoder concatenates the processed data of image encoder and text encoder to merge them into final representation as state.

\paragraph{Module-specific search spaces.}

Based on the state encoder architecture with multiple modules of distinct functions and specific architecture types in \Cref{fig:composite_game}, we define module-specific search spaces, as detailed in \Cref{tab:search_space_game}. Following \citet{chitty2022neural}, key hyperparameters commonly used in NAS for CNNs with pooling layers are included: the type of pooling method (denoted as "pooling type" in the table), the layer where the CNN is followed with the pooling layer ("pooling layer"), the type of activation function ("activation"), the size of kernels ("kernel size"), the number of channels ("channel number") and number of neural network layers ("depth") for the image encoder. Following \citet{chitty2022neural}, key parameters commonly used in NAS for RNNs with embedding layer are included: the dimension of embedding layer ("dimension"), the hidden size of RNNs ("hidden size"), the ratio of drop-out in RNNs ("dropout") and number of neural network layers ("depth") for the text encoder. Following \citet{chitty2022neural}, key variance of fusion method could be summarized as: the merge type ("merge type"), the dimension of fusion network ("dimension"), the type of activation function ("activation") and the hidden size of networks ("hidden size"). Common value ranges are set for all search parameters, resulting in a total of approximately 19 million possible architectures throughout the search space.

\begin{table}[htbp]
\centering
\caption{Module-specific search spaces for the MiniGrid benchmark where bold values denote the configurations used by the Expert baseline.}
\label{tab:search_space_game}

\begin{subtable}[t]{.65\linewidth}
\centering
\caption{Image Encoder (CNN \& Pooling)}
\begin{tabular}{ll}
\toprule
\textbf{Hyperparameter} & \textbf{Value Range} \\
\midrule
Pooling type   & \{\textbf{max}, average\} \\
Pooling layer  & \{\textbf{1}, ..., depth\} \\
Activation     & \{\textbf{sigmoid}, tanh, relu, elu\} \\
Kernel size    & \{1, \textbf{2}\} \\
Channel number & \{8, \textbf{16}, 32, 64\} \\
Depth          & \{2, \textbf{3}, 4, 5\} \\
\bottomrule
\end{tabular}
\end{subtable}%

\bigskip
\begin{subtable}[t]{0.45\linewidth}
\centering
\caption{Text Encoder (Embedding \& GRU)}
\begin{tabular}{ll}
\toprule
\textbf{Hyperparameter} & \textbf{Value Range} \\
\midrule
Dimension   & \{16, \textbf{32}, 64, 128\} \\
Hidden size & \{64, \textbf{128}, 256, 512\} \\
Dropout     & \{\textbf{0.0}, 0.1, 0.2, 0.3\} \\
Depth       & \{\textbf{1}, 2, 3, 4\} \\
\bottomrule
\end{tabular}
\end{subtable}%
\hfill
\begin{subtable}[t]{0.51\linewidth}
\centering
\caption{Fusion Encoder (Fusion Module)}
\begin{tabular}{ll}
\toprule
\textbf{Hyperparameter} & \textbf{Value Range} \\
\midrule
Merge type  & \{\textbf{cat}, add, gate, bilinear\} \\
Dimension   & \{\textbf{128}, 256, 512, 1024\} \\
Activation  & \{\textbf{sigmoid}, tanh, relu, elu\} \\
Hidden size & \{32, \textbf{64}, 128, 256\} \\
\bottomrule
\end{tabular}
\end{subtable}

\end{table}

\paragraph{Baseline alignment.}

Similar with \Cref{apdx:baseline_alignment}, when applying traditional NAS methods to NAS for state encoders in RL for goal-oriented tasks, certain corresponding mappings are required: (i) Performance: The accuracy used in ENAS and PEPNAS is replaced here by average return; similarly, the gradient in DARTS, which reflects validation performance, is also mapped to average return. (ii) Sample size: In PEPNAS, the sample size for validating candidates within each generation increases incrementally, which corresponds here to an incremental increase in training steps when validating candidates within each generation.

\paragraph{RL training and evaluation details.}

\begin{figure}[htbp]
    \centering
    \begin{subfigure}[b]{0.49\linewidth}
        \centering
        \includegraphics[width=\linewidth]{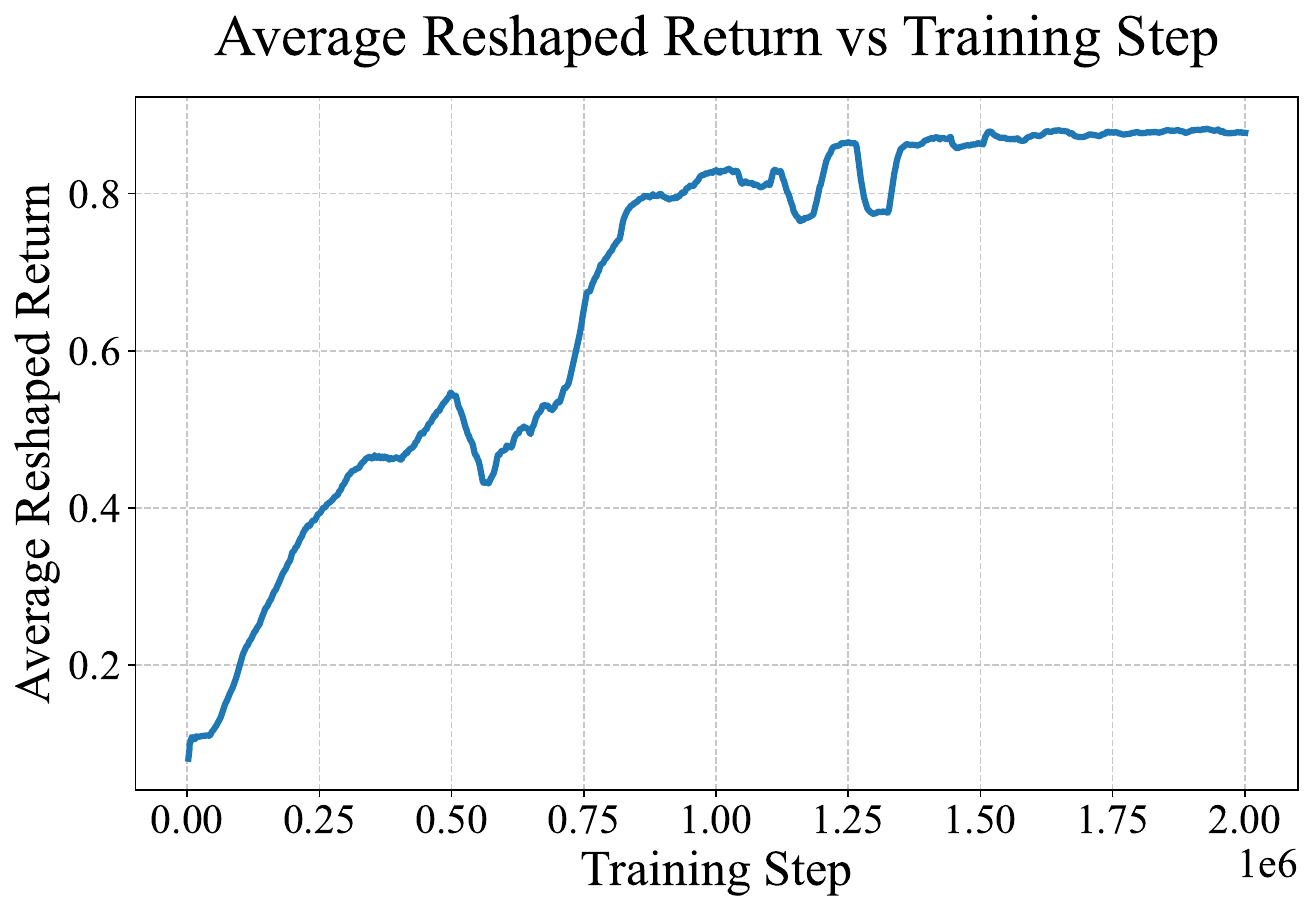} 
        \caption{Reshaped return over 2M training steps}
        \label{fig:rR_2M}
    \end{subfigure}
    \hfill
        \begin{subfigure}[b]{0.49\linewidth}
        \centering
        \includegraphics[width=\linewidth]{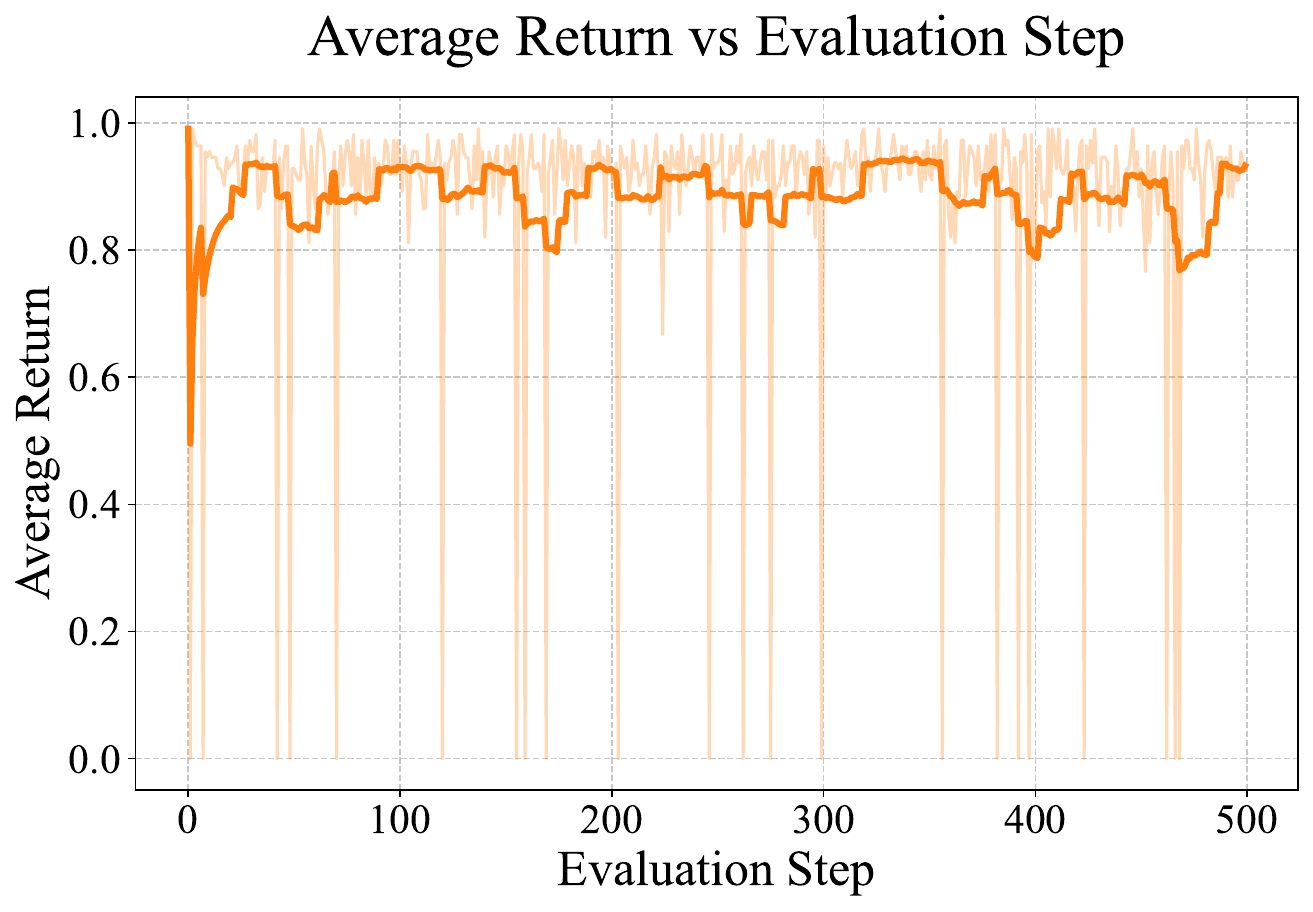} 
        \caption{Average return over 500 evaluation steps}
        \label{fig:R_500}
    \end{subfigure}
    \hfill
    \begin{subfigure}[b]{0.49\linewidth}
        \centering
        \includegraphics[width=\linewidth]{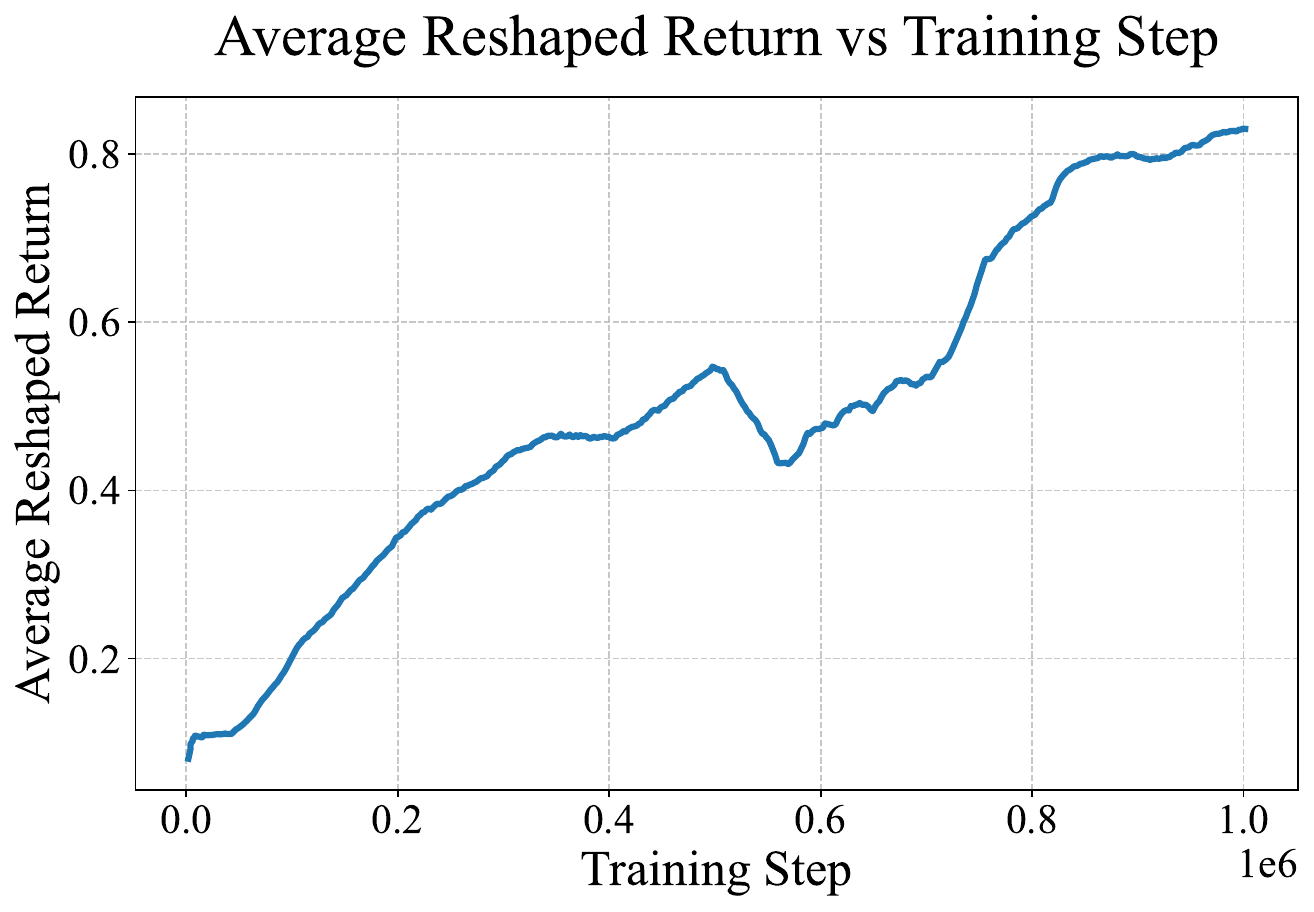} 
        \caption{Reshaped return over 1M training steps}
        \label{fig:rR_1M}
    \end{subfigure}
    \begin{subfigure}[b]{0.49\linewidth}
        \centering
        \includegraphics[width=\linewidth]{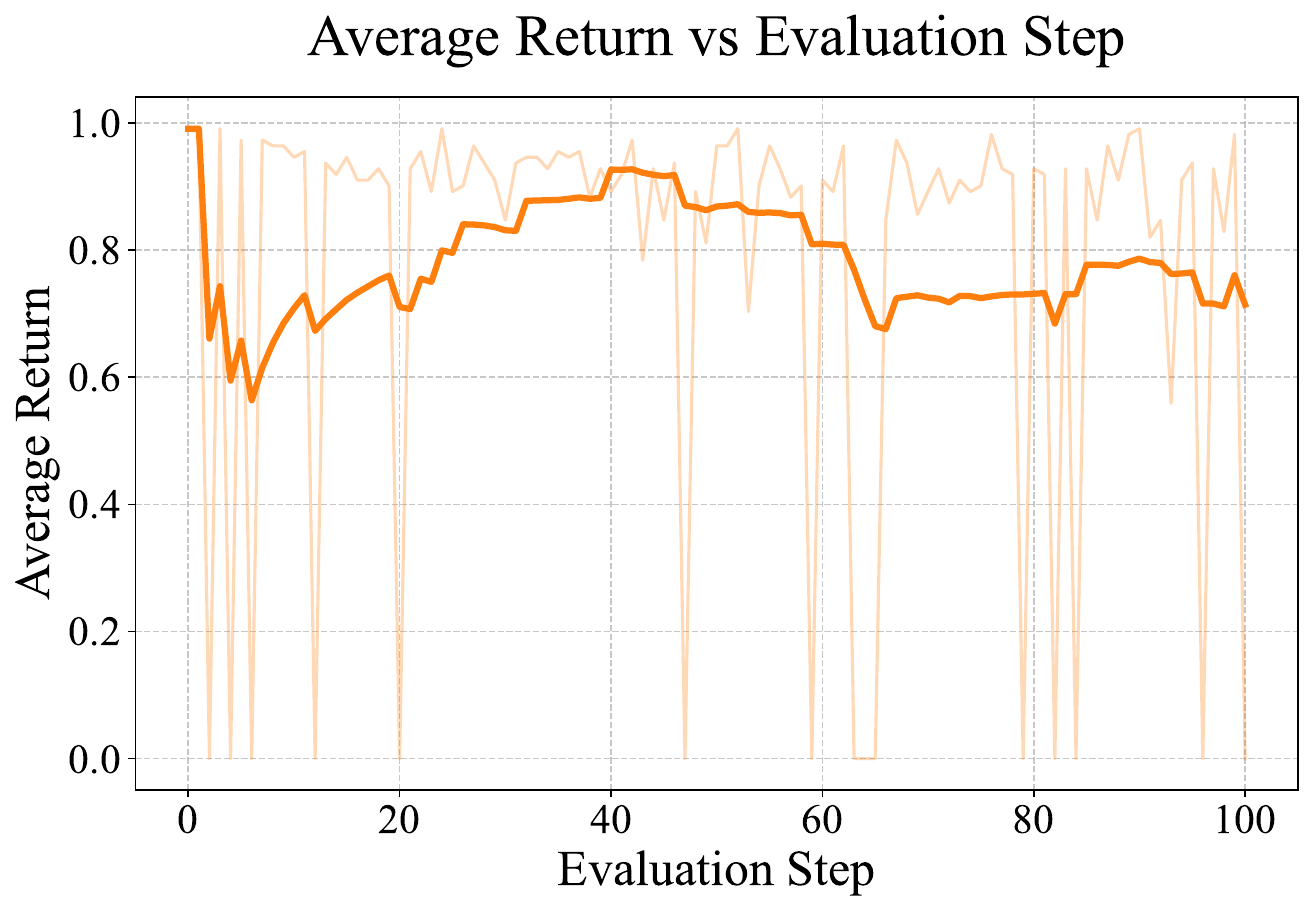} 
        \caption{Average return over 100 evaluation steps}
        \label{fig:R_100}
    \end{subfigure}
    \caption{Reshaped return and during training and average return during evaluation in the MiniGrid goal-oriented tasks.}
    \label{fig:rl_training_analysis_game}
\end{figure}

RL training requires sufficient steps for policy convergence, typically manifested by reward. In the MiniGrid goal-oriented tasks, agent uses reshaped return to train policy and requires several trials to finish the task under the trained policy. Thus, RL policy evaluation also requires adequate steps to encompass multiple such trials. To balance RL training convergence, evaluation comprehensiveness, and the cost of evaluating state encoder architectures, we analyzed the average reshaped return of training (recorded every 2,048 steps over 2M steps of baseline RL training) and average return of evaluation (recorded every step over 500 steps of baseline evaluation) as shown in \Cref{fig:rl_training_analysis_game}. The results indicate that the average reshaped return converges around 1M steps, while the average return converges around 100 steps. Based on this, each candidate state encoder architecture was evaluated by integrating it into the RL framework for 1M steps of training and 100 steps of evaluation.

\subsection{ManiSkill robotic control}

\paragraph{Architecture of the state encoder.}

\begin{figure}[htp]
    \centering
    \includegraphics[width=12cm]{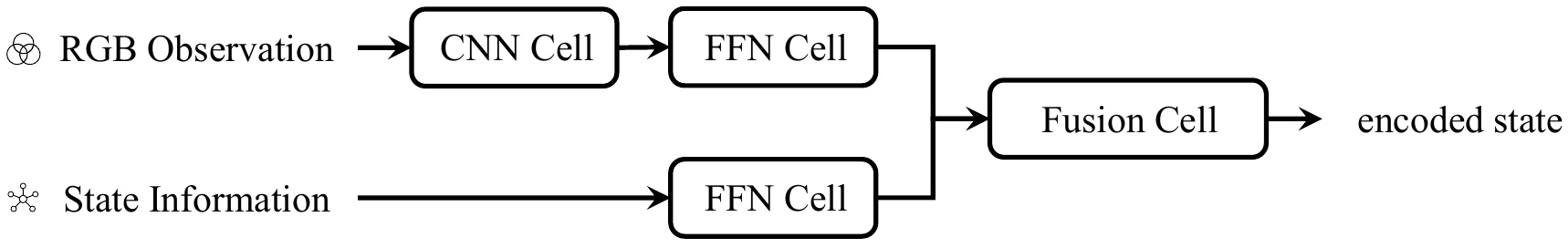} 
    \caption{Composite architecture of the state encoder in the ManiSkill robotic control}
    \label{fig:composite_robotic}
\end{figure}

Following \citet{tao2024maniskill3}, the architecture of the
state encoder includes an RGB encoder and a information encoder for respective data encoding while the fusion encoder processes their concatenated outputs to generate the encoded state for RL training, as illustrated in \Cref{fig:composite_robotic}.  The RGB encoder uses CNNs added with FFNs to process RGB observations. The information encoder uses FFNs to represent other state information. The fusion encoder concatenates the processed data of image encoder and text encoder to merge them into final representation as state.

\section{Additional Experiment Results}
\label{apdx:additional_results}

\paragraph{Ablation studies}

\begin{figure}[htp]
    \centering
    \includegraphics[width=0.6\textwidth]{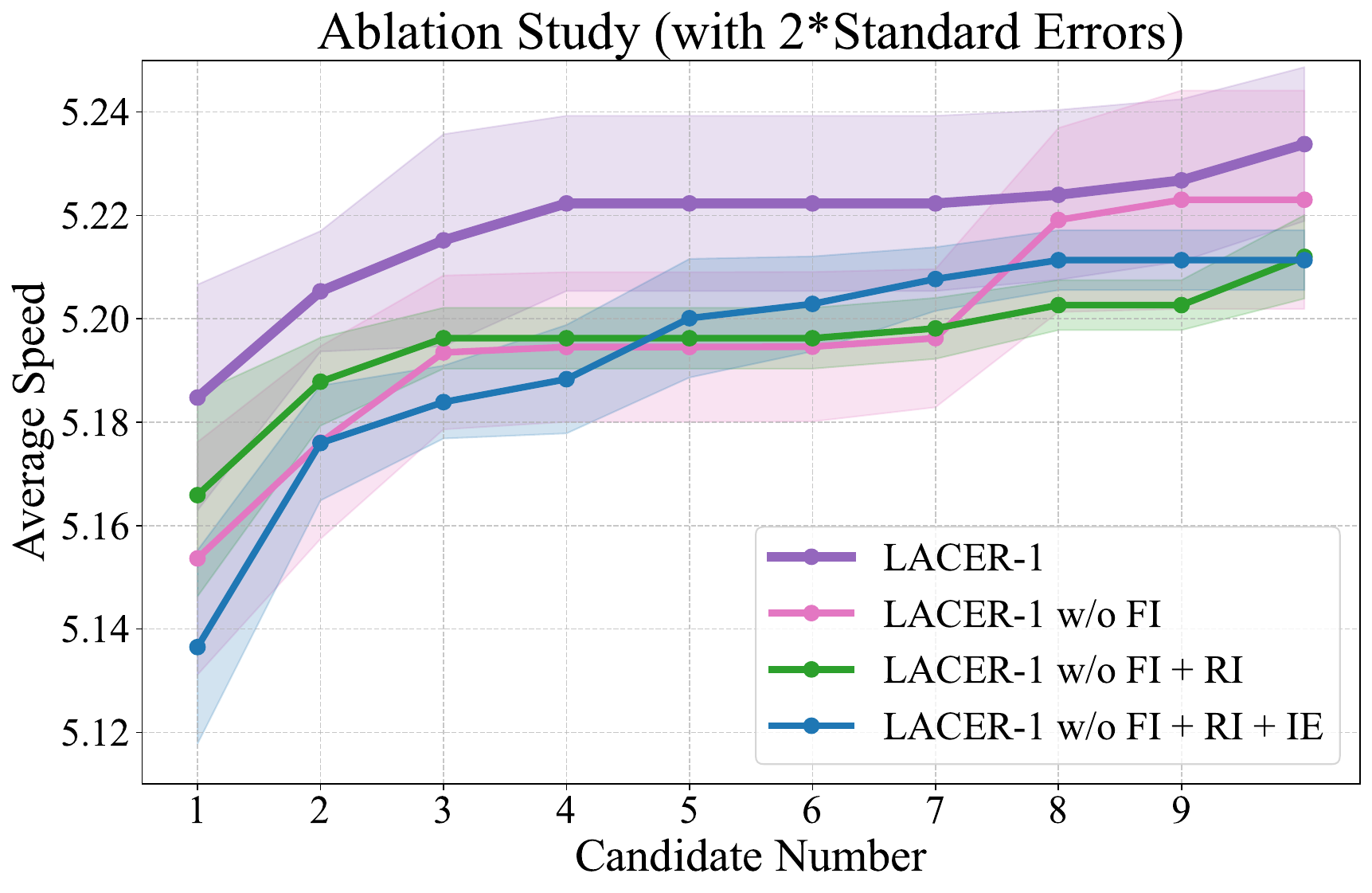} 
    \caption{Performance comparison of LACER-1 with and without different key prompt components.}
    \label{fig:ablation}
\end{figure}

To verify the necessity of each core module in the proposed LACER method, we designed a series of ablation experiments, in which three key prompt components were removed, respectively: the feature information (FI), the average reward (RI), and the initial architecture evaluation (IE). This resulted in three variant methods: LACER-1 without FI, LACER-1 without FI + RI and LACER-1 without FI + RI + IE. The results in \cref{fig:ablation} indicate that the original LACER-1 method achieves the best performance. When any of the components mentioned above is removed, the performance of LACER deteriorates significantly. This phenomenon demonstrates that each of these components in our LACER method is useful and essential, and their collaborative operation contributes to the superior performance of the proposed method.

\paragraph{Time cost analysis}

\begin{figure}[t!]
    \centering  
    \begin{subfigure}[b]{0.48\textwidth}  
        \centering  
        \includegraphics[width=\textwidth]{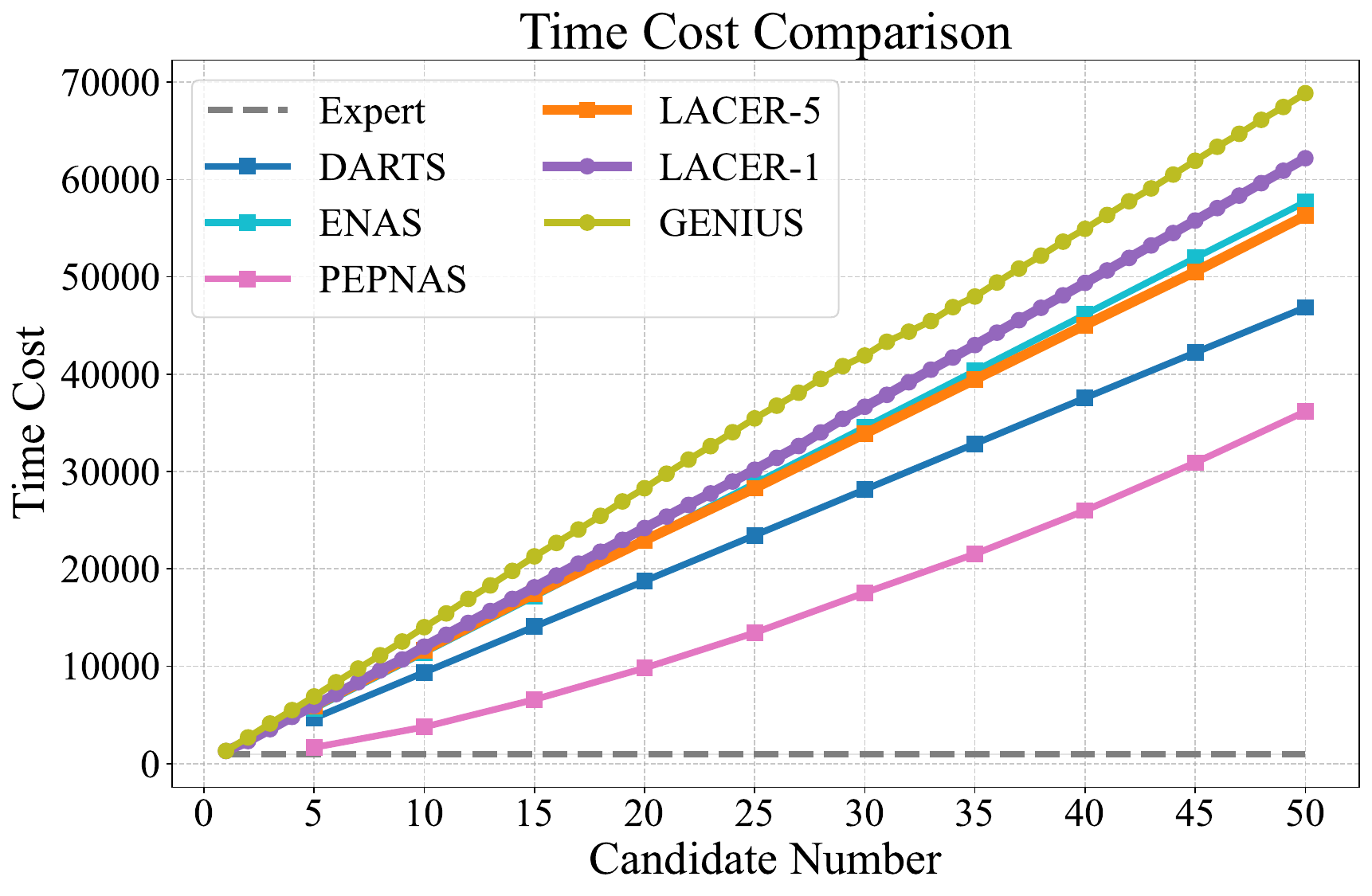}  
        \caption{Time cost comparison} 
        \label{subfig:cumulative_time}  
    \end{subfigure}
    \hfill  
    \begin{subfigure}[b]{0.48\textwidth}
        \centering
        \includegraphics[width=\textwidth]{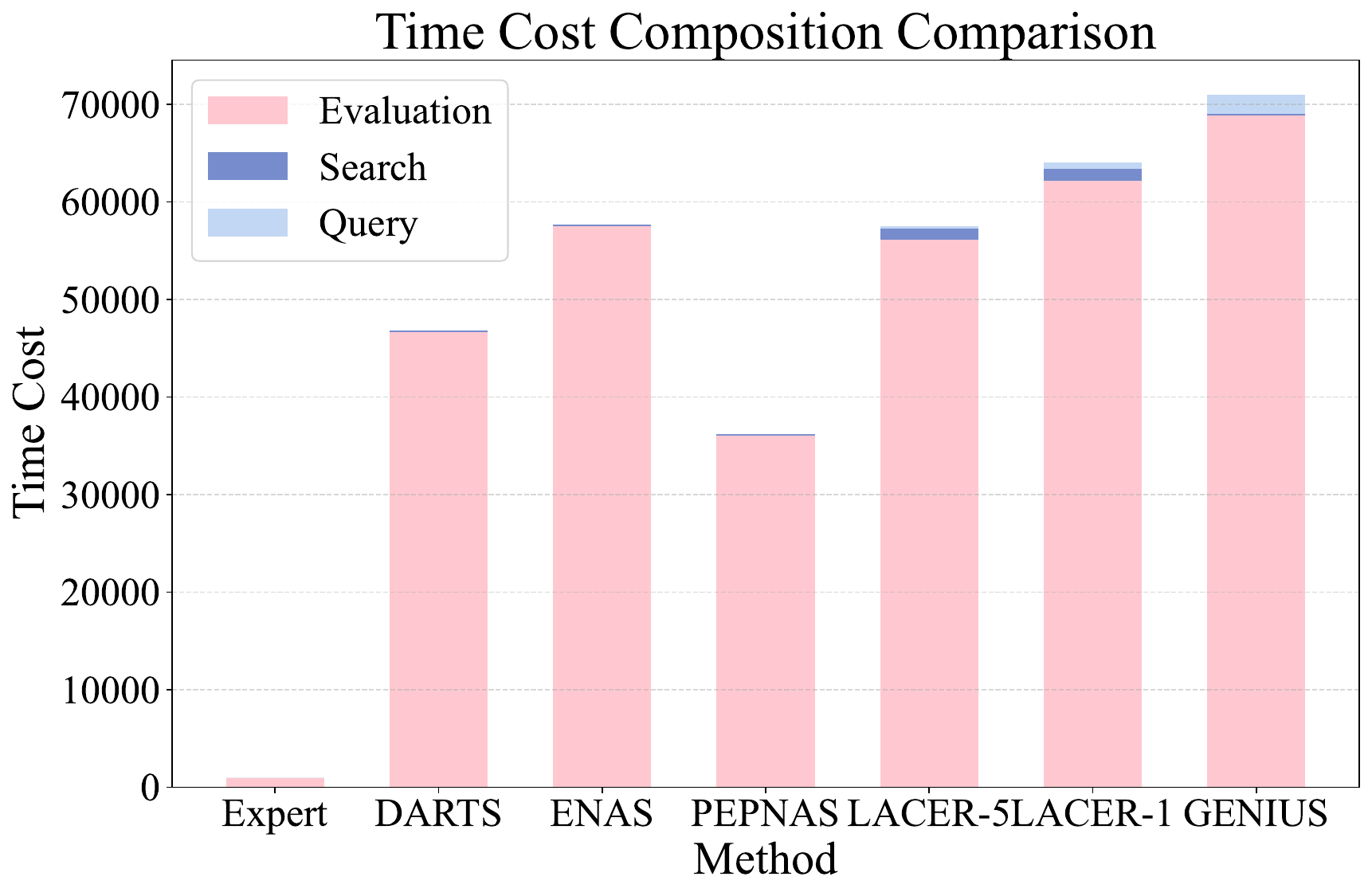}
        \caption{Time cost composition comparison}  
        \label{subfig:time_composition}  
    \end{subfigure}
    \caption{Left: Time cost comparison between our two LACER variants and baselines; Right: Time cost composition comparison between our two LACER variants and baselines. The query time accounts for 1\% of the overall time cost of LACER, while the evaluation time accounts for over 97\% of the overall time cost of all methods.}
    \label{fig:time_cost}  
\end{figure}

The time cost of traditional NAS methods is composed of the evaluation time of each candidate and the search time of each generation, while the search cost of LLM-based NAS methods includes the query time of LLM and the processing time of analyzing results of candidates and constructing prompts. To analyze the impact of query time, we run LACER and other baselines on an Intel Core i7 CPU and an NIDIA RTX 4070 GPU to avoid the time cost of node scheduling. We then compare the time cost of the above compositions as shown in \Cref{fig:time_cost}, where evaluation refers to the evaluation time, search refers to the search time that does not include query time, and query refers to the query time. The result indicates that the query time of LLM has a negligible impact on the overall time cost compared to the evaluation time.

\end{document}